\documentclass[pmlr]{jmlr}% new name PMLR (Proceedings of Machine Learning)

\RequirePackage{graphicx}
 \usepackage{booktabs}
\usepackage{longtable}% for long tables
\usepackage{makecell}
\usepackage{tcolorbox}
\usepackage{float}
\makeatletter
\def\set@curr@file#1{\def\@curr@file{#1}} %temp workaround for 2019 latex release
\makeatother
\usepackage[load-configurations=version-1]{siunitx} % newer version

\theorembodyfont{\upshape}
\theoremheaderfont{\scshape}
\theorempostheader{:}
\theoremsep{\newline}

\title[Shape Over Intensity]{Shape Over Intensity: Directional Topological Encoding for False Positive Reduction in Intracranial Aneurysm Detection}

\author{\Name{Akshay Gokhale}
       \Email{akshay.gokhale22@spit.ac.in}\\ 
       \addr Computer Science and Engineering\\
       Sardar Patel Institute of Technology\\
       Mumbai, India
       \AND
       \Name{Mansi Dhamne}
       \Email{mansi.dhamne22@spit.ac.in}\\ 
       \addr Computer Engineering\\
       Sardar Patel Institute of Technology\\
       Mumbai, India
}

\begin{document}

\maketitle

\begin{abstract}
Automated detection of intracranial aneurysms (IAs) from CT angiography (CTA) is severely hindered by high false-positive rates. Traditional convolutional neural networks (CNNs) rely fundamentally on local pixel intensities, causing them to systematically confuse true saccular aneurysms with healthy vascular bifurcations. This geometric ambiguity is especially catastrophic for small, clinically critical lesions ($<$3 mm), where standard detection sensitivity often plummets below 60\%.

To resolve this, we propose a plug-and-play, topology-aware false-positive reduction framework. We evaluate the Smooth Euler Characteristic Transform (SECT)—a directional mathematical representation that encodes global 3D vascular geometry independently of intensity—against standard persistence-based summaries (Persistence Images and Landscapes). The framework is rigorously stress-tested on a curated, heavily stratified subset of the multi-institutional RSNA 2025 dataset, explicitly designed to challenge models with anatomically plausible bifurcation mimics.

SECT demonstrates exceptional, classifier-agnostic discriminative power, achieving an AUC of 0.943, substantially outperforming direction-agnostic persistence methods ($\text{AUC} \sim 0.68$). Crucially, our topological filter exhibits a clinical performance inversion: it excels on the most challenging sub-3 mm cohort, maintaining a 0.943 AUC and 78.5\% sensitivity even under a strict 95\% specificity constraint. Furthermore, the representation proves highly resilient to hardware-specific artifacts, maintaining a 0.927 mean AUC under strict leave-one-scanner-out (LOGO) validation across four distinct manufacturers.

By explicitly capturing asymmetric geometric invariants rather than intensity profiles, directional topological representations reliably resolve the primary structural confounder in IA detection. This establishes SECT as a highly robust, scanner-agnostic downstream filter ready for integration into hybrid deep-learning diagnostic pipelines.

\end{abstract}

\section{Introduction}
The automated detection of Intracranial Aneurysms (IAs) remains a critical challenge in machine learning for healthcare. IAs are abnormal, localized dilations of cerebral arteries affecting approximately 3\% to 7\% of the adult population \citep{Vlak2011PrevalenceOU, joo2025methodological}. If left untreated, these weakened vessel walls can rupture, causing a fatal subarachnoid hemorrhage. While Digital Subtraction Angiography (DSA) remains the gold standard, Computed Tomography Angiography (CTA) is the primary non-invasive screening modality used in clinical practice \citep{hsu2025survey}. However, reliably identifying IAs on CTA is notoriously difficult even for skilled radiologists. This difficulty is magnified for small lesions ($<3$ mm in diameter), where sensitivity frequently drops to between 64\% and 74\% \citep{bizjak2023systematic}. Developing an automated diagnostic aid is paramount for early intervention, yet current deep learning systems generate an overwhelming number of false alarms, severely hindering their clinical viability.

Addressing this problem is challenging due to the inherent geometric limitations of standard convolutional neural networks (CNNs). CTA natively suffers from lower spatial resolution and low vessel-to-background contrast, making small vascular structures visually indistinct because of surrounding tissue or noise. More fundamentally, modern CNNs rely heavily on local pixel intensities and texture statistics. Because a saccular aneurysm and a healthy vascular bifurcation exhibit nearly identical local intensity distributions in a CTA volume, CNNs consistently confuse the two. This geometric ambiguity---treating a complex branching intersection exactly like a bulging sac---translates into a prohibitive number of false positives per scan. Standard intensity-based CNNs do not explicitly model global structural invariants; they cannot mathematically differentiate the asymmetric, spherical protrusion of an aneurysm dome from the symmetric branching of a healthy vessel bifurcation. Consequently, previous deep learning architectures, such as 3D U-Nets and ResNets, often exhibit lesion-level sensitivities for small aneurysms of merely 56\% and fail to generalize across different scanner acquisition protocols \citep{bizjak2023systematic, joo2025methodological}. 

In this work, we introduce a topologically-aware false-positive reduction framework that explicitly models global structural invariants to differentiate aneurysms from complex vessel bifurcations. Rather than replacing existing deep learning architectures, our module is designed as a plug-and-play false-positive reduction filter intended to sit downstream of high-sensitivity CNN candidate generators. Drawing on Persistent Homology (PH) from Topological Data Analysis (TDA), we capture coordinate-independent geometric descriptors that track the topological evolution of a shape across multiple scales. However, we hypothesize that standard direction-agnostic topological summaries are insufficient for cerebrovascular structures. Instead, we evaluate the Smooth Euler Characteristic Transform (SECT), a directional topological representation that captures the asymmetric spatial configuration of aneurysms relative to their parent vessels. We validate our framework on a curated, heavily stratified subset of the multi-institutional RSNA 2025 dataset \citep{rsna2025_aneurysm}, explicitly designed to stress-test geometric discrimination against anatomically plausible bifurcation mimics. \footnote{To support reproducibility, the full codebase can be made available on request.}

% Our core contributions are as follows:
% \begin{itemize}
%     \item \textbf{Directional Topological Superiority:} We demonstrate that the directional geometry captured by SECT ($0.94$ AUC) vastly outperforms standard direction-agnostic persistence representations like Persistence Images and Landscapes ($\sim0.68$ AUC).
%     \item \textbf{Sub-3mm Performance Inversion:} Remarkably, our topological filter exhibits a performance inversion relative to standard deep learning models: it achieves its highest discriminative power (AUC $> 0.94$) on the smallest, most clinically challenging aneurysms ($<3$ mm). In this sub-3 mm regime, our framework retains a $0.7845$ sensitivity even under strict 95\% specificity constraints.
%     \item \textbf{Inter-Scanner Generalization:} We demonstrate exceptional cross-domain robustness. Under strict leave-one-scanner-out (LOGO) validation across four distinct manufacturers, SECT achieves a highly stable mean AUC of $0.9273$, proving resilience against hardware-specific intensity shifts.
% \end{itemize}

\begin{figure*}[t]
\centering
\includegraphics[width=\textwidth]{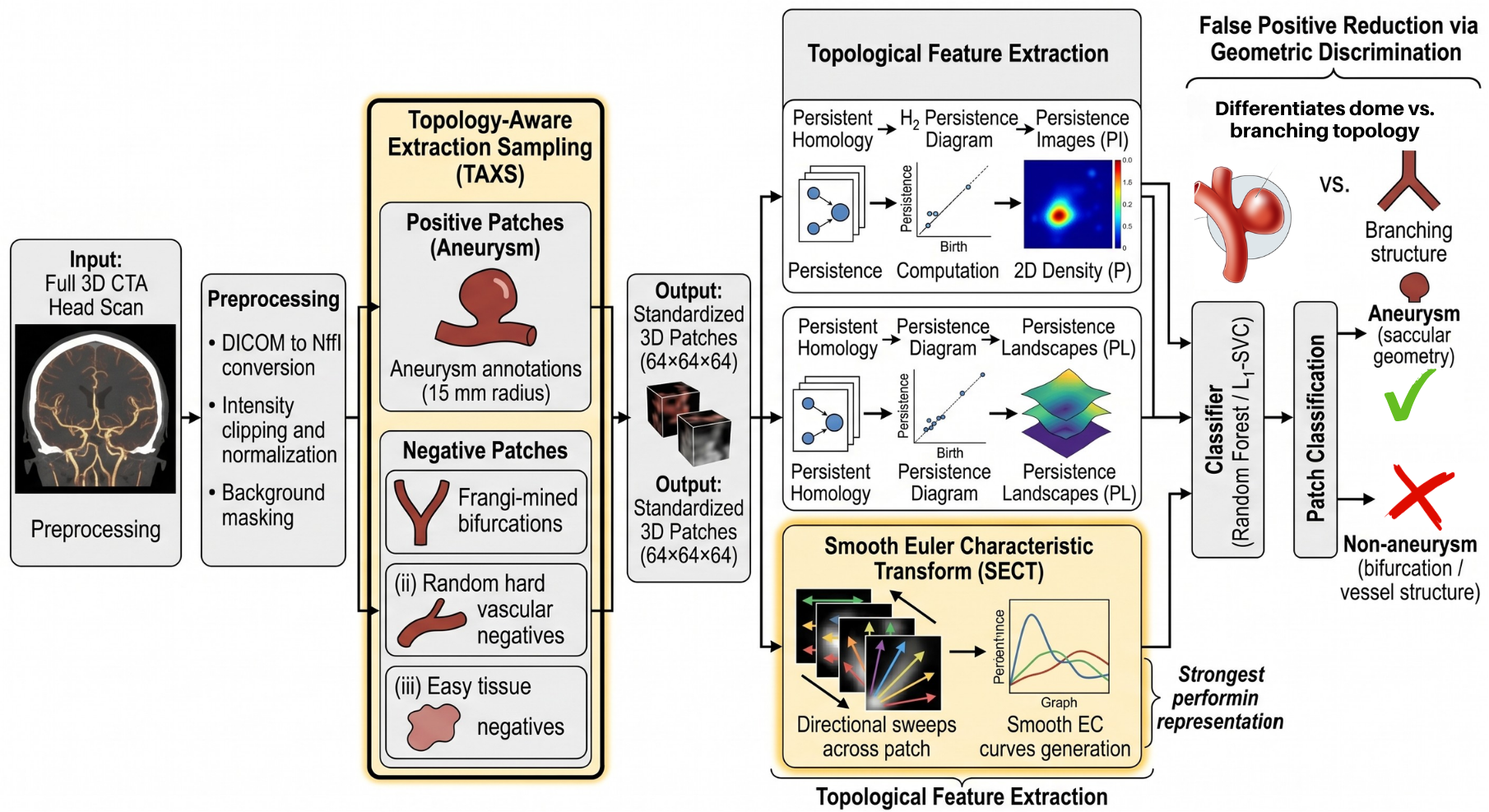}
\caption{\textbf{Topology-Aware False Positive Reduction Pipeline.} Input CTA volumes undergo intensity standardization before localized 3D patches are extracted via the Topology-Aware Extraction Sampling (TAXS) strategy. Patches are heavily stratified into positive aneurysms and highly confounding negative classes (e.g., Frangi-mined bifurcations). Topological features are extracted via Persistent Homology (PI, PL) and directional filtration (SECT). Acting as a downstream filter, SECT provides the strongest representation, enabling the classifier to accurately discriminate saccular geometry from complex branching structures based purely on shape rather than pixel intensity.}
\label{fig:pipeline_overview}
\end{figure*}

\subsection*{Generalizable Insights about Machine Learning in the Context of Healthcare}
Our work highlights a critical lesson for healthcare machine learning: models relying exclusively on local visual patterns (e.g., intensity and texture) are inherently brittle in clinical scenarios where healthy and pathological structures share similar density profiles. We demonstrate that incorporating explicit mathematical representations of global shape and spatial organization can resolve this geometric ambiguity in a principled manner. Furthermore, by focusing on topological invariants rather than raw pixel values, we show that structural properties remain stable across varied imaging conditions. This provides a pathway to improving generalizability across disparate hospital systems and scanner manufacturers without requiring exhaustive retraining, emphasizing that integrating clinical intuition about anatomy with geometric machine learning leads to more robust, trustworthy, and deployable diagnostic systems.

\section{Related Work}

\subsection{Deep Learning for Intracranial Aneurysm Detection}
Automated intracranial aneurysm (IA) detection has predominantly been approached as a segmentation or object detection task using architectures such as 3D U-Nets, ResNets, and GLIA-Net \citep{bo2021toward,bizjak2023systematic,hsu2025survey}. While these systems achieve high sensitivity for large aneurysms, they fundamentally rely on localized pixel intensities and texture statistics. Consequently, they falter on small lesions ($<3$ mm), where sensitivity plummets to roughly 56\%, and produce an overwhelming number of false positives (FPs) by confusing saccular aneurysms with healthy vascular bifurcations \citep{bizjak2023systematic}. Furthermore, as noted in recent scoping reviews, most studies lack external validation, multi-institutional data, and size-stratified metrics, yielding optimistic aggregate performance that proves brittle in clinical deployment \citep{joo2025methodological}.

\subsection{Geometry-Aware and FP-Reduction Methods}
To address the high FP rates of pure intensity-based CNNs, modern computer-aided detection (CAD) systems frequently adopt two-stage pipelines: a high-sensitivity candidate generator followed by a dedicated FP-reduction classifier. This paradigm has proven highly effective in both pulmonary nodule and aneurysm detection architectures \citep{Setio_Traverso_Bel_Berens_Bogaard_Cerello_Chen_Dou_Fantacci_Geurts_etal._2017}. Within these secondary filters, explicitly modeling vascular geometry is critical. Techniques ranging from classical multi-scale vesselness filters (e.g., Frangi) to modern deep-learning skeletonization are frequently employed to extract centerlines and suppress tubular structures \citep{frangi1998multiscale}. More recently, Graph Neural Networks (GNNs) have been applied directly to extracted vascular trees to enable complex bifurcation modeling and artery-vein classification \citep{vos2024graph, hampe2024graph}. However, these graph- and centerline-based approaches rely heavily on accurate preliminary segmentation, which is highly error-prone in the low-contrast, noisy regime of small aneurysms.

\subsection{Topological Data Analysis in Medical Imaging}
To overcome the limitations of localized texture classifiers and brittle skeletonization algorithms, Topological Data Analysis (TDA) offers a mathematically rigorous framework for extracting robust global geometries \citep{singh2023topological, singh2025unraveling}. TDA investigates the underlying shape of a scalar field through algebraic topology, tracking the presence of connected components ($H_0$), loops ($H_1$), and voids ($H_2$) across multiple scales. The core mechanism, Persistent Homology (PH) \citep{otter2017roadmap, clough2019explicit}, provides mathematical summaries that are inherently invariant to deformations and the localized scanner noise prevalent in CTA. While TDA has been successfully applied to characterize general tumor morphology and pulmonary tissue by separating meaningful anatomical structures from transient noise \citep{Boehm_Fink_Attenberger_Becker_Behr_Reiser_2008, crawford2020predicting}, its application in distinguishing nuanced cerebrovascular structures remains largely unexplored.

\subsection{Vectorization of Topological Features for ML Integration}
Integrating topological priors into deep learning pipelines presents a fundamental computational challenge. The native output of PH is a Persistence Diagram (PD)---a variable-cardinality multi-set of birth-death coordinates that cannot be directly ingested by standard classifiers \citep{adams2017persistence, som2020pi}. To bridge this gap, methods such as Persistence Images (PI) utilize lifetime-weighted Gaussian kernels to project PDs into stable Euclidean grids \citep{chepushtanova2015persistence, adams2017persistence}. Alternatively, Persistence Landscapes (PL) encode topological lifespans via layered, integrable functions mapped to Banach spaces \citep{bubenik2015statistical}. However, PI and PL often struggle to reliably capture the subtle $H_2$ topological void signatures necessary for identifying sub-3 mm aneurysms, and standard PH calculations suffer from an $O(N^3)$ computational bottleneck on high-resolution 3D medical volumes. To circumvent this, recent advances introduced the Smooth Euler Characteristic Transform (SECT) \citep{Turner_Mukherjee_Boyer_2014, crawford2020predicting}. SECT generates smooth, continuous vector curves based on directional sensitivity to asymmetric shape deformations, bypassing the traditional bottleneck of PH entirely. 

Despite advancements in theoretical vectorization, existing methods have not systematically utilized explicit topological priors to resolve the geometric ambiguity between intracranial aneurysms and vessel bifurcations. By transforming complex vascular geometries into ML-compatible topological descriptors (PIs, PLs, and SECT vectors), our methodology explicitly leverages these representations to act as a definitive, shape-driven downstream filter for current CNN-based detection systems.

\section{Methodology}
\subsection{Problem Formulation}

We formalize intracranial aneurysm (IA) detection as a localized binary classification problem on 3D patches extracted from CT angiography (CTA) volumes, rather than a full-volume detection or segmentation task. Prior work and empirical observations indicate that aneurysms and vessel bifurcations exhibit similar intensity and texture characteristics in CTA, leading to high false positive rates in CNN-based systems. 

The distinction between these structures is therefore hypothesized to be geometric rather than intensity-based, motivating representations that capture intrinsic shape properties. Thus, as illustrated in the classification stage of Figure \ref{fig:pipeline_overview}, our problem revolves around discriminating aneurysm patches from anatomically similar vessel bifurcations at the patch level, allowing us to focus explicitly on clinically challenging small aneurysms ($<3$ mm).

\subsection{Dataset}
\label{sec:dataset}
All experiments are conducted on the RSNA Intracranial Aneurysm Detection Challenge 2025 dataset (Kaggle), a large-scale, multi-institutional collection of neuroimaging studies \citep{rsna2025_aneurysm}. This dataset includes heterogeneous acquisitions across multiple scanner manufacturers and imaging protocols, providing a clinically realistic benchmark for evaluating model generalization.

For the scope of this work, we restrict our analysis exclusively to Computed Tomography Angiography (CTA) volumes. This restriction is motivated by CTA's clinical prevalence as a primary, non-invasive screening modality. Furthermore, CTA presents a particularly challenging computational setting due to its lower spatial resolution and reduced sensitivity for small aneurysms compared to the Digital Subtraction Angiography (DSA) gold standard.

Leaving out the corrupted files, the accessible CTA cohort comprises $\approx$1808 unique scans. Within this subset, 973 patients have atleast one expert-annotated intracranial aneurysm, localized via spatial coordinates and specific slice instance identifiers.

The final dataset consists of 1,201 aneurysm-positive patches and a heavily stratified negative class designed to test geometric discrimination. The negative class includes 3,576 Frangi-mined hard negatives (anatomically plausible vessel bifurcations), 1,787 random hard negatives (other dense vascular structures), and 1,787 easy negatives (background tissue). The exact algorithmic extraction and class balancing protocols are detailed in Section \ref{app:patch_construction}.

To ensure strict data hygiene and prevent data leakage, dataset splitting is strictly enforced at the patient level. Each aneurysm contributes at most one positive patch, and no patient appears in both the training and evaluation splits. Patients with multiple aneurysms may therefore contribute multiple positive patches, but all patches from a given patient are assigned to the same split.

\subsection{Patch Construction and Pre-processing}
To ensure geometric consistency and scanner invariance across multi-institutional acquisitions, raw DICOM series are first converted to volumetric NIfTI format while strictly preserving native scanner coordinate frames. Prior to patch extraction, volumes undergo strict intensity standardization: we apply anatomical window clipping and threshold-based masking to remove background air and bone artifacts, followed by per-volume Z-score normalization on the valid voxels.

Following volumetric standardization, we implement a Topology-Aware Extraction Sampling (TAXS) strategy, explicitly designed to emulate the false-positive distribution of a high-sensitivity candidate generation network. Positive patches are anchored to expert-annotated aneurysm centroids with a 15 mm physical radius and interpolated to a uniform $64 \times 64 \times 64$ voxel grid. To rigorously evaluate geometric discrimination, negative patches are stratified into three categories:
\textit{(i) Frangi-mined hard negatives}: Extracted using a multi-scale Frangi vesselness filter \citep{frangi1998multiscale} to specifically isolate anatomically plausible, complex vascular bifurcations, \textit{(ii) Random hard negatives}: High-intensity local maxima geometrically distant from true lesions, \textit{(iii) Easy tissue negatives}: Sampled from intermediate-intensity background regions. (See Appendix \ref{alg:patch_construction} for detailed algorithmic extraction)

Sampling is performed at the patient level to maintain a controlled negative-to-positive ratio while preventing data leakage. The dataset is explicitly constructed to target a key failure mode of existing systems—confusion between aneurysms and vascular bifurcations—by including anatomically plausible hard negatives. It is further enriched with small ($<$3 mm) aneurysms, enabling focused evaluation on clinically challenging cases where detection performance is most critical.

\subsection{Topological Feature Extraction and Vectorization}
Persistent Homology (PH) provides a principled framework for capturing multi-scale topological structure by tracking the birth and death of features across a continuous filtration \citep{zomorodian2005topology, edelsbrunner2014short}. For a 3D scalar field, such as a CTA patch, PH summarizes the evolution of connected components ($H_0$), loops ($H_1$), and voids ($H_2$) as the intensity threshold varies. The native output of PH is a persistence diagram (PD), a variable-cardinality multi-set of birth–death coordinates, $PD = \{(b_i, d_i)\}_{i=1}^k$, where each point encodes the lifespan of a specific topological feature.

While PDs provide a complete topological summary, their non-Euclidean nature and varying set sizes render them incompatible with standard machine learning pipelines. Consequently, multiple vectorization methods have been proposed to embed PDs into fixed-dimensional spaces while preserving stability under perturbations \citep{adams2017persistence, carriere2020perslay}. Among these, Persistence Images (PI) and Persistence Landscapes (PL) are widely adopted due to their theoretical guarantees and compatibility with conventional classifiers \citep{adams2017persistence, bubenik2020persistence}. However, both representations share a fundamental limitation: they produce direction-agnostic summaries. They encode \emph{which} topological features exist and their relative persistence, but discard critical information regarding \emph{how these features are spatially organized within the volume}.

This limitation is critical for intracranial aneurysm detection. Aneurysms and natural vessel bifurcations frequently exhibit similar local topological invariants, but differ fundamentally in their geometric configuration relative to the parent vessel. Capturing this asymmetry requires representations that preserve spatial context. To address this, we evaluate three complementary vectorization strategies: PI and PL as canonical baselines, and the Smooth Euler Characteristic Transform (SECT) \citep{crawford2020predicting}—a directional topological representation explicitly designed to encode geometric asymmetry.

\subsubsection{Persistent Homology Setup}
\label{sec:ph-setup}
For each preprocessed patch $\tilde{P}_c$, we construct a cubical complex using the voxel grid as the underlying topological domain. Persistent homology is computed using the \texttt{GUDHI} library (v3.11.0) \citep{maria2014gudhi}. To ensure that bright vascular structures enter the filtration first, we define a superlevel filtration over the normalized scalar intensity field $f: \tilde{P}_c \rightarrow [0,1]$. This is implemented by evaluating the sublevel sets of the negated field:
\begin{equation}
    \mathcal{F}_\alpha = {x \in \tilde{P}_c \mid -f(x) \leq \alpha}.
\end{equation}

We specifically isolate the second homology group ($H_2$). Rather than representing literal enclosed cavities---as aneurysms are open structures---$H_2$ features in our superlevel filtration encode the 3D intensity gradient topology. They act as morphological signatures of localized convexity, distinguishing the sac-like geometry of aneurysm domes from the tubular branching of bifurcations. We empirically validate this representational capability using parameterized synthetic phantoms, demonstrating robust separation of saccular and branching structures across varying scales and noise regimes (see Appendix \ref{app:phantom_validation}).

For each clinical patch, we extract the $H_2$ persistence diagram, $PD_2(f)$. To suppress background noise and transient imaging artifacts, we apply a strict persistence threshold, retaining only finite features with a lifespan $(d_i - b_i) > \varepsilon$, where $\varepsilon = 0.15$ in normalized intensity units. This parameter aggressively filters noise while preserving the subtle, short-lived topological signatures characteristic of sub-3~mm aneurysms. The resulting sparse diagrams serve as direct inputs for the PI and PL vectorizations.

\subsubsection{Persistence Images}
\label{sec:pi}
Persistence Images (PI) \citep{adams2017persistence} provide a stable, finite-dimensional vector representation of persistence diagrams, enabling seamless integration with conventional learning algorithms. To construct a PI, each point in the persistence diagram is first transformed from birth–death coordinates to birth–persistence coordinates:
\begin{equation}
(b_i, d_i) \mapsto (b_i, p_i), \quad \text{where } p_i = d_i - b_i.
\end{equation}

This transformation isolates the lifespan of a feature ($p_i$) from its spatial emergence ($b_i$). Each point then contributes a smooth Gaussian kernel centered at its new location, generating a continuous persistence surface:
\begin{equation}
\rho(x, y) = \sum_{i=1}^{k} w(b_i, p_i), \phi_{\sigma}(x - b_i,, y - p_i),
\end{equation}
where $\phi_{\sigma}$ is a Gaussian kernel with bandwidth $\sigma$, and $w(b_i, p_i)$ is a specialized weighting function. $w$ is explicitly designed to reduce the effects of low-persistence features near the diagonal, effectively suppressing topological noise while modulating the contribution of each valid feature.

Finally, this continuous persistence surface $\rho(x,y)$ is discretized over a fixed, uniform grid $\Omega \subset \mathbb{R}^2$ to produce the persistence image. Each pixel value corresponds to the two-dimensional integral of $\rho$ over that specific grid cell:
\begin{equation}
PI(u,v) = \iint_{\text{cell}_{u,v}} \rho(x,y), dx, dy.
\end{equation}
The resulting image matrix is flattened into a fixed-length vector for downstream classification.

\subsubsection{Persistence Landscapes}
\label{sec:pl}
Persistence Landscapes (PL) \citep{bubenik2015statistical} address the non-Euclidean limitations of persistence diagrams by mapping them into a functional representation within a Banach space (completely normalized vector space), allowing the direct application of classical statistical tools.

Given a persistence diagram $PD = \{(b_i, d_i)\}_{i=1}^{n}$, each birth–death pair is first associated with a tent function:
\begin{equation}
f_{(b_i, d_i)}(t) =
\begin{cases}
0, & t \notin (b_i, d_i), \\
t - b_i, & t \in (b_i, \tfrac{b_i + d_i}{2}], \\
d_i - t, & t \in (\tfrac{b_i + d_i}{2}, d_i).
\end{cases}
\end{equation}
Each tent function forms an isosceles triangle with a peak height proportional to the feature's persistence, centered at its topological midpoint $(b_i + d_i)/2$. 

The persistence landscape is then formally defined as a sequence of functions $\lambda_k: \mathbb{R} \rightarrow \mathbb{R}$, where $\lambda_k(t)$ is the $k$-th largest value of this set of tent functions:
\begin{equation}
    \lambda_k(t) = \text{$k$-th largest value of } \{ f_{(b_i,d_i)}(t) \}_{i=1}^{n}.
\end{equation}

The resulting landscape can be interpreted as a stacked sequence of continuous curves derived from barcode intervals, where $\lambda_1$ captures the most dominant topological features and higher-order landscapes encode progressively weaker, noisier signals.

\subsubsection{Smooth Euler Characteristic Transform (SECT)}
\label{sec:sect}
The Smooth Euler Characteristic Transform (SECT) \citep{crawford2020predicting} provides a directional topological representation that maps each patch to a collection of smooth Euler characteristic curves. Unlike PI and PL, which summarize topology without preserving directional organization, SECT encodes directional geometric structure of vascular shapes rather than relying on local intensity patterns. This is supported by controlled synthetic experiments (Appendix~\ref{app:phantom_validation}), which demonstrate that SECT consistently distinguishes saccular and bifurcation geometries under varying noise, scale, and orientation. 

The SECT builds upon the Euler characteristic (EC), a fundamental topological invariant that compactly encodes the number of connected components, loops, and higher-dimensional voids within a shape. To capture multiscale directional structure, SECT computes EC curves by tracking the topological evolution of the shape across sublevel set filtrations, defined by height functions along uniformly sampled directions on a sphere.

Because raw EC curves are piecewise constant, integer-valued functions containing sharp discontinuities, they must be smoothed. The curves are mean-centered and integrated to produce continuous, piecewise-linear functional representations known as Smooth Euler Characteristic (SEC) curves. This transformation ensures the representation resides in a well-behaved Hilbert space ($\mathbb{L}^2$) suitable for statistical learning. The formal mathematical derivations of the EC and the smoothing transformations are detailed in Appendix \ref{app:sect}.

Operationally, SECT is computed by first restricting the normalized patch to a vascular foreground mask, sweeping this mask along a uniformly sampled set of directions, and concatenating the discretized, smoothed EC curves into a fixed-dimensional feature vector. We utilize a vascular threshold $\tau_v = 0.40$ to define the foreground mask. This is conceptually distinct from the persistence threshold $\varepsilon = 0.15$ used in Section \ref{sec:ph-setup}: $\tau_v$ determines the raw intensity support of the directional shape, whereas $\varepsilon$ filters short-lived topological artifacts from persistence diagrams. The final SECT representation uses $D=128$ directions, $T=25$ filtration steps, Gaussian smoothing along the filtration axis, and $\ell_2$ normalization. A procedural overview is provided in Appendix~\ref{alg:sect}.

\section{Experiments and Results}
% classification results --> scanner robustness --> size stratified results --> finegrained analysis of performance on negative class samples --> pi/pl stability --> topological seperabilty
\subsection{Experimental Setup and Cohort Characterization}
All methods operate on the same set of CTA patches described in Section~\ref{sec:dataset}. Pre-processing is fixed across all experiments, including intensity clipping, Gaussian smoothing ($\sigma = 0.5$), and normalization to $[0,1]$. Prior to downstream classification, we empirically verified that persistent homology natively captures a meaningful geometric signal; pairwise bottleneck distance analysis confirms a stable topological hierarchy distinguishing aneurysms from Frangi-mined bifurcations (see Appendix \ref{app:topo_seperability}).

Hyperparameters for each representation are selected via a preliminary grid search and then held fixed for all downstream evaluations to isolate representation quality from tuning effects. Sensitivity analysis (Appendix~\ref{app:sect_params}) shows that performance is largely invariant to reasonable variations in key parameters, indicating that the chosen configurations lie in a stable regime.
 
\subsection{Classification Benchmark Across Topological Representations}
\label{sec:classification_benchmark}

We evaluate the discriminative capacity, stability, and computational efficiency of three topological vectorization methods: Persistence Images (PI), Persistence Landscapes (PL), and the Smooth Euler Characteristic Transform (SECT). Each method maps the 3D geometry of CTA patches into a fixed-dimensional feature vector, enabling evaluation under a shared classification pipeline (Section~\ref{sec:sect}).

To assess whether performance gains are primarily \emph{representation-driven} rather than classifier-dependent, we evaluate each method using two distinct classifiers: a non-linear Random Forest (RF) and an L1-regularized Linear Support Vector Classifier (L1-SVC). We report four complementary metrics: Area Under the ROC Curve (AUC) reported as mean $\pm$ standard deviation under 5-fold cross-validation, empirical Lipschitz constant $L$ (measuring sensitivity to controlled Gaussian noise), feature dimensionality, and average end-to-end inference time per patch (including feature extraction and classification).

Both PI and PL achieve moderate discrimination performance, with AUC values clustered around $0.68$, and exhibit sensitivity to classifier choice as relative rankings are not consistently preserved. In contrast, SECT achieves substantially higher performance under both Random Forest (0.9433 AUC) and L1-SVC (0.9312 AUC), indicating that its advantage is representation-driven rather than classifier-dependent.

\begin{figure*}[t]
\centering
\includegraphics[width=\textwidth]{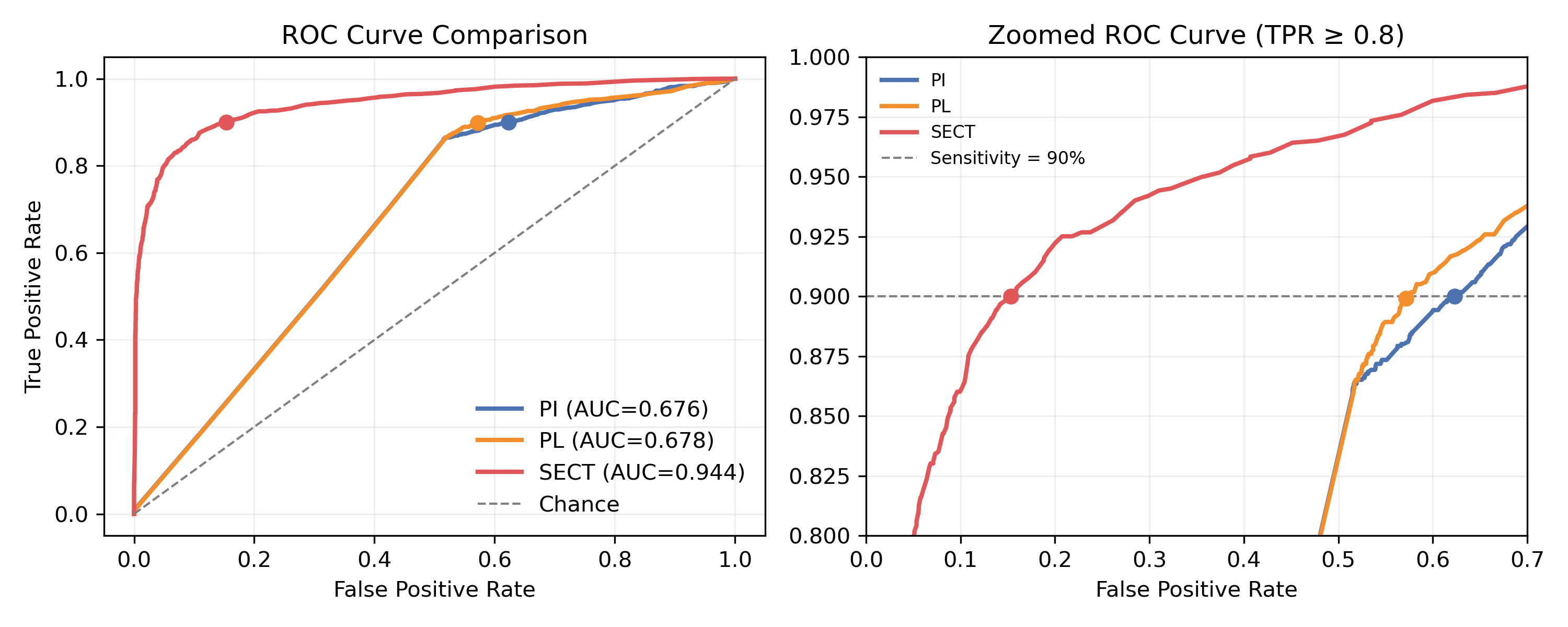}
\caption{
ROC comparison of PI, PL, and SECT using 5-fold out-of-fold predictions. 
Right: zoomed view of the high-sensitivity operating region. 
Markers indicate operating points at 90\% sensitivity. 
SECT consistently achieves higher sensitivity at lower false positive rates, demonstrating improved discrimination of aneurysms from vascular structures.
}
\label{fig:roc_comparison}
\end{figure*}

Figure~\ref{fig:roc_comparison} presents ROC curves along with a zoomed view of the high-sensitivity operating region. SECT consistently dominates PI and PL across the ROC space, with particularly strong separation in the high-sensitivity regime. At clinically relevant operating points (e.g., 90\% sensitivity), SECT achieves substantially lower false positive rates compared to persistence-based baselines. Detailed operating-point analysis, including FPR at fixed sensitivity and bifurcation-specific false positive rates, is provided in Appendix~\ref{app:fp_analysis}.

Regarding robustness, PI exhibits high sensitivity to input perturbations (large empirical Lipschitz constant, $L$), whereas PL yields a smoother but less discriminative embedding. SECT occupies an optimal intermediate regime, maintaining bounded sensitivity without sacrificing discriminative power. Because SECT bypasses intermediate persistence diagrams, its stability is measured end-to-end directly against physical image noise ($\sigma$) rather than the Wasserstein diagram distance used for PI and PL. Although this difference in metric spaces prevents direct numerical comparison of $L$ values (detailed in Appendix~\ref{app:lipschitz_methodology}), SECT's formulation offers a more clinically relevant measure of robustness to actual scanner variance. Table~\ref{tab:main_results} summarizes these findings, highlighting SECT's consistent performance advantage.

% Given the substantial performance gains of SECT, a higher dimensionality and computational cost is an acceptable tradeoff for offline candidate ranking and false-positive reduction settings.

% \begin{table*}[t]
% \centering
% \small
% \begin{tabular}{lcccccc}
% \toprule
% \textbf{Method} 
% & \textbf{Dim} 
% & \textbf{L1-SVC AUC (↑)} 
% & \textbf{RF AUC (↑)} 
% & \textbf{Native Domain for $L$} 
% & \textbf{Empirical $L$ (95\% CI)} 
% & \textbf{Inference Time (s)} \\
% \midrule
% PI  
% & 900 
% & 0.6791 $\pm$ 0.0066 
% & 0.6745 $\pm$ 0.0077 
% & $W_p$ (Diagram Space)
% & 430.5 [418.6, 444.6]
% & ~1 sec \\

% PL  
% & 750 
% & 0.6515 $\pm$ 0.0086 
% & 0.6809 $\pm$ 0.0075 
% & $W_p$ (Diagram Space)
% & 3.9 [1.9, 5.9] 
% & ~1 sec \\

% SECT  
% & 3200 
% & \textbf{0.9312 $\pm$ 0.0083} 
% & \textbf{0.9433 $\pm$ 0.0076} 
% & $\sigma$ (Image Space)
% & 40.3 [28.9, 133.9]
% & ~11 sec \\
% \bottomrule
% \end{tabular}

% \caption{
% Comparison of topological representations across classifiers, stability, and computational cost. 
% Performance is reported as mean AUC $\pm$ standard deviation under 5-fold cross-validation. 
% SECT consistently outperforms persistence-based representations (PI, PL) across both classifiers, 
% while exhibiting moderate stability and higher computational cost due to increased feature dimensionality. Computational efficiency is reported as average end-to-end inference time per patch, including both feature extraction and classification.
% }
% \label{tab:main_results}
% \end{table*}

\begin{table*}[t]
\centering
\footnotesize
\setlength{\tabcolsep}{4pt} % Slightly reduces horizontal padding between columns
\caption{Comparison of topological representations across classifiers, stability, and computational cost. 
Performance is reported as mean AUC $\pm$ standard deviation under 5-fold cross-validation. 
SECT consistently outperforms persistence-based representations (PI, PL) across both classifiers, 
while exhibiting moderate stability and higher computational cost due to increased feature dimensionality. Computational efficiency is reported as average end-to-end inference time per patch, including both feature extraction and classification. ($W_p$ stands for Diagram Space and $\sigma$ for Image Space}
\begin{tabular}{lcccccc}
\toprule
\textbf{Method} 
& \textbf{Dim} 
& \makecell{\textbf{L1-SVC} \\ \textbf{AUC ($\uparrow$)}} 
& \makecell{\textbf{RF} \\ \textbf{AUC ($\uparrow$)}} 
& \makecell{\textbf{Native Domain} \\ \textbf{for $L$}} 
& \makecell{\textbf{Empirical $L$} \\ \textbf{(95\% CI)}} 
& \makecell{\textbf{Inference} \\ \textbf{Time (s)}} \\
\midrule
PI  
& 900 
& 0.6791 $\pm$ 0.0066 
& 0.6745 $\pm$ 0.0077 
& $W_p$ 
& 430.5 [418.6, 444.6]
& $\sim$1 \\

PL  
& 750 
& 0.6515 $\pm$ 0.0086 
& 0.6809 $\pm$ 0.0075 
& $W_p$
& 3.9 [1.9, 5.9] 
& $\sim$1 \\

SECT  
& 3200 
& \textbf{0.9312 $\pm$ 0.0083} 
& \textbf{0.9433 $\pm$ 0.0076} 
& $\sigma$
& 40.3 [28.9, 133.9]
& $\sim$11 \\
\bottomrule
\end{tabular}
\label{tab:main_results}
\end{table*}

\subsection{Size-Stratified Evaluation of SECT}

While overall classification metrics provide a global measure of discriminative capacity, clinical utility critically depends on performance across varying lesion sizes. Because small aneurysms ($< 3$ mm) constitute the primary failure mode of automated CNN pipelines \citep{bizjak2023systematic}, we evaluate whether SECT maintains its discriminative power across standard aneurysm size strata. This stratification isolates the true geometric properties of individual lesions and prevents overall performance metrics from being artificially skewed by easily detectable large aneurysms.

Positive patches are partitioned into three clinically standard categories (Small, Medium, and Large). Size annotations are extracted dynamically at the patch level to ensure spatial fidelity (detailed in Appendix \ref{app:size_verification}). For each stratum, we construct a balanced evaluation task by pairing the size-specific positive patches against the shared pool of Frangi-mined hard negatives. We report the Area Under the ROC Curve (AUC) alongside sensitivity measured at strict clinical operating points (95\% and 99\% specificity) using the Random Forest classifier.

\begin{table}[h]
\centering
\caption{Size-stratified SECT performance. Performance is reported as mean AUC $\pm$ standard deviation. Sensitivity is evaluated at fixed specificity (Sp.) thresholds to simulate strict false-positive reduction constraints.}
\begin{tabular}{lcccc}
\toprule
\textbf{Size Stratum} & \textbf{$n$} & \textbf{AUC ($\uparrow$)} & \textbf{Sens @ 95\% Sp.} & \textbf{Sens @ 99\% Sp.} \\
\midrule
Small ($< 3$ mm)  & 942 & \textbf{0.9434 $\pm$ 0.0100} & \textbf{0.7845} & \textbf{0.6306} \\
Medium (3--7 mm) & 112 & 0.8634 $\pm$ 0.0209 & 0.5714 & 0.3036 \\
Large ($> 7$ mm)  & 147 & 0.9178 $\pm$ 0.0147 & 0.6395 & 0.3946 \\
\midrule
Overall& 1201& 0.9405 $\pm$ 0.0095 & 0.7918 & 0.6003 \\
\bottomrule
\end{tabular}
\label{tab:size_results}
\end{table}

\begin{figure}[h!]
\centering
\includegraphics[width=\linewidth]{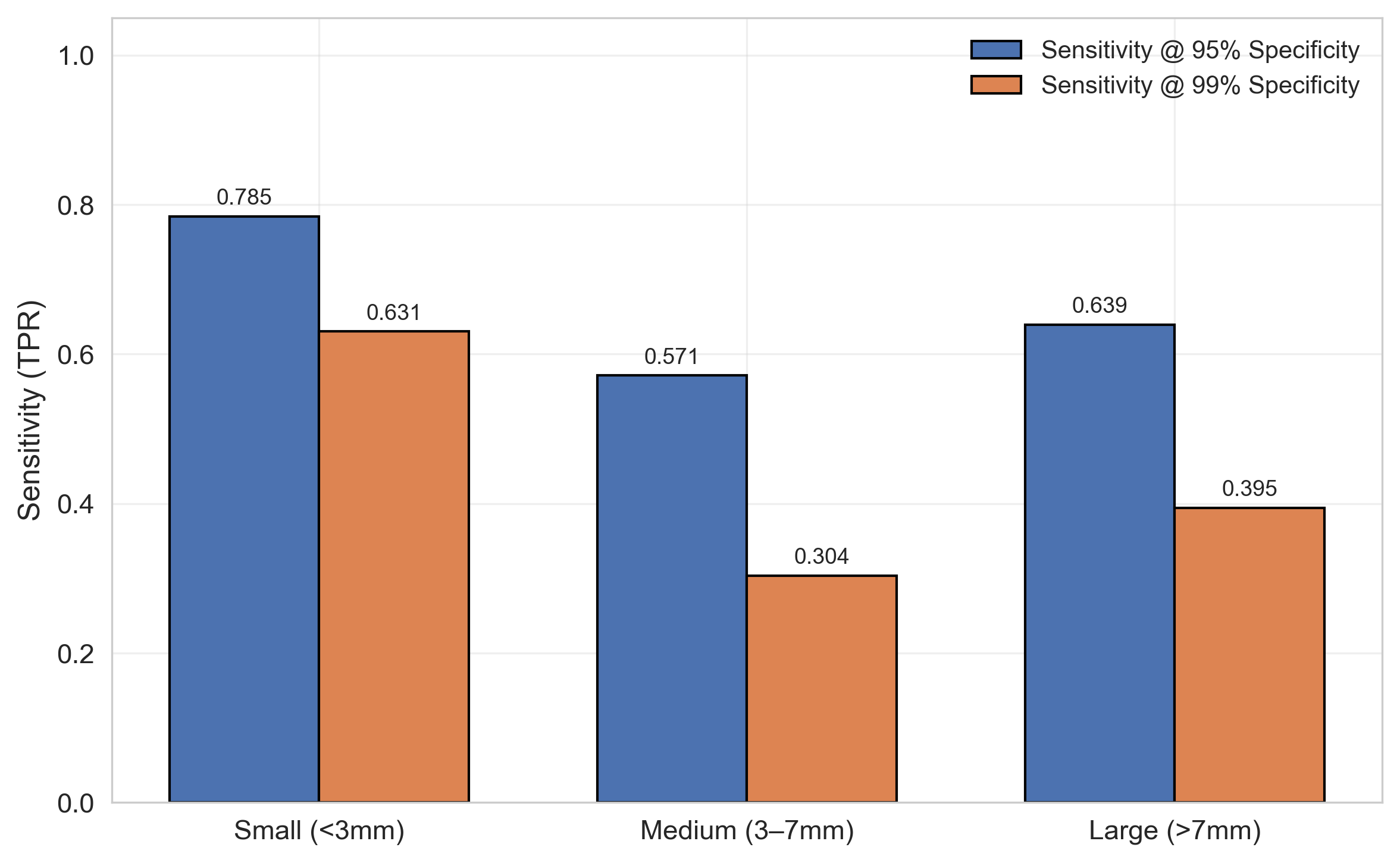}
\caption{\textbf{Sensitivity across aneurysm size strata.} SECT maintains robust detection capabilities for clinically challenging small aneurysms, even under strict specificity constraints, bypassing the typical performance degradation observed in intensity-based CNNs.}
\label{fig:size_stratified_plot}
\end{figure}

As detailed in Table \ref{tab:size_results} and visualized in Figure \ref{fig:size_stratified_plot}, SECT exhibits exceptional performance on the sub-3 mm cohort, achieving an AUC of $0.9434 \pm 0.0100$. Contrary to the characteristic sensitivity collapse observed in traditional intensity-based networks, SECT maintains a remarkable sensitivity of 0.7845 for small aneurysms even when forced to operate at a strict 95\% specificity. Even at an extreme 99\% specificity constraint, the method retains a 0.6306 detection rate.

Interestingly, while overall performance remains strong, sensitivity degrades for the Medium and Large strata at stricter operating points. We attribute this variance to two primary factors. First, the severe natural class imbalance in our dataset yields significantly smaller sample sizes for these cohorts ($n=112$ and $n=147$, respectively), reducing statistical stability. Second, this degradation highlights a geometric constraint of our extraction pipeline: the fixed 15 mm physical patch radius may inadvertently truncate the topological boundaries of large aneurysms. Additionally, medium-sized aneurysms often exhibit transitional geometries that more closely mimic the smooth tubular branching of complex bifurcations. 

% Future iterations of this pipeline may resolve this geometric ambiguity by utilizing multi-scale, adaptive patch resolutions.

\subsection{Inter-Scanner Robustness of SECT}
A primary barrier to the clinical translation of automated intracranial aneurysm detection is the severe performance degradation models experience across different scanner manufacturers and acquisition protocols. Traditional CNNs frequently overfit to scanner-specific intensity profiles and noise calibrations. We therefore evaluate whether the geometric signatures captured by SECT preserve discriminative performance across diverse acquisition domains.

Positive and negative patches are assigned scanner labels derived from native DICOM metadata (Siemens, GE, Toshiba/Canon, and Philips). We systematically evaluate scanner robustness under two distinct paradigms: within-scanner evaluation and leave-one-scanner-out (LOGO) validation. For the within-scanner baseline, we report stratified 5-fold cross-validation AUC using exclusively the patches originating from a single manufacturer. For the LOGO evaluation, we enforce strict domain separation: the model is trained on patches from all manufacturers \textit{except} scanner $S$, and evaluated exclusively on the held-out scanner $S$. This prevents domain leakage and yields a definitive test of cross-scanner generalization.

\begin{table}[h]
\centering
\caption{Cross-scanner robustness evaluation. The model demonstrates exceptional generalization, with LOGO performance matching or even exceeding within-scanner baselines.}
\begin{tabular}{l c c | c c}
\toprule
& & \textbf{Within-Scanner} & \multicolumn{2}{c}{\textbf{Leave-One-Scanner-Out (LOGO)}} \\
\cmidrule(lr){3-3} \cmidrule(lr){4-5}
\textbf{Scanner} & \textbf{$n$} & \textbf{AUC ($\uparrow$)} & \textbf{AUC ($\uparrow$)} & \textbf{AP ($\uparrow$)} \\
\midrule
Siemens           & 377 & 0.9510 $\pm$ 0.0199 & 0.9369 & 0.9052 \\
GE                & 333 & 0.9529 $\pm$ 0.0071 & \textbf{0.9585} & 0.8897 \\
Toshiba/Canon     & 287 & 0.9011 $\pm$ 0.0178 & 0.8904 & 0.8240 \\
Philips           & 204 & 0.8963 $\pm$ 0.0092 & \textbf{0.9233} & 0.8889 \\
% \midrule
% \textit{Mean}     & - & \textit{0.9253} & \textit{0.9273} & \textit{0.8769} \\
\bottomrule
\end{tabular}
\label{tab:scanner_robustness}
\end{table}

As detailed in Table \ref{tab:scanner_robustness}, SECT achieves consistently high discriminative performance within individual acquisition domains, with within-scanner AUCs ranging from 0.8963 to 0.9529. More importantly, under the strict, leakage-free LOGO evaluation, SECT maintains highly robust performance across all held-out cohorts (mean AUC $0.9273 \pm 0.0247$).

Supporting Principal Component Analysis (Appendix \ref{app:scanner_analysis}) reveals that while minor scanner-associated variance exists within the high-dimensional SECT feature space, the distributions across the four manufacturers remain substantially intertwined without forming isolated clusters. Consequently, while the representation is not mathematically perfectly scanner-invariant, the learned topological decision boundary transfers highly effectively across acquisition domains, proving SECT's resilience to the inter-scanner calibration shifts that typically confound intensity-based networks.

\subsection{Generalizability Under Mixed Negative Distributions}
We evaluate whether the topological representations remain discriminative under shifts in the negative class distribution, moving beyond the Frangi-only evaluation toward clinically realistic candidate mixtures. In clinical practice, an automated detection pipeline will generate a heterogeneous pool of candidates, encompassing vessel bifurcations, complex vascular tortuosity, and non-vascular background tissue artifacts.

We categorize negative patches into three groups: Frangi (F) bifurcation candidates, Hard (H) non-bifurcating vascular structures, and Easy (E)non-vascular background tissue. Our primary evaluation utilizes a clinically motivated mixture of 5:3:2 (F:H:E). To probe sensitivity to distributional variation, we evaluate additional configurations while keeping the positive cohort and the overall negative-to-positive ratio constant. Classification utilizes the Random Forest pipeline established in Section \ref{sec:classification_benchmark}.

\begin{table}[h]
\centering
\caption{SECT performance under varying negative class compositions.}
\begin{tabular}{l c c}
\toprule
\textbf{Mixture Setting} & \textbf{Ratio (F:H:E)} & \textbf{AUC ($\uparrow$)} \\
\midrule
Frangi-only       & 1:0:0 & 0.940 \\
Hard-dominant     & 3:5:2 & 0.926 \\
Primary clinical  & 5:3:2 & 0.922 \\
Uniform           & 1:1:1 & 0.914 \\
Easy-dominant     & 4:2:4 & 0.905 \\
\bottomrule
\end{tabular}
\label{tab:mixed_all}
\end{table}

As reported in Table \ref{tab:mixed_all}, SECT maintains robust discriminative capacity across all mixture configurations. Under the primary clinical mixture (5:3:2), SECT achieves an AUC of AUC of $0.922$ and an Average Precision of $0.844$. This represents a minor, acceptable degradation from the idealized Frangi-only setting (AUC $0.940$), indicating that the representation does not collapse when exposed to realistic background noise.

To understand the specific failure modes within this mixed distribution, we conducted a per-class false-positive analysis on the primary clinical mixture (Table \ref{tab:per_class_fpr}).

\begin{table}[h]
\centering
\caption{Per-class false-positive rates (FPR) under the primary 5:3:2 mixed setting.}
\begin{tabular}{l c c c}
\toprule
\textbf{Negative Type} & \textbf{$n$} & \textbf{FPR ($\downarrow$)} & \textbf{Mean Pred. Score} \\
\midrule
Hard Vascular  & 1073 & \textbf{0.003} & 0.078 \\
Frangi Bifurcations & 1788 & 0.024 & 0.113 \\
Easy Non-Vascular   & 715  & 0.122 & 0.219 \\
\bottomrule
\end{tabular}
\label{tab:per_class_fpr}
\end{table}

The breakdown reveals a highly illuminating, counter-intuitive result. Standard vascular structures (Hard) are almost perfectly rejected (FPR = 0.003), and Frangi-mined bifurcations are suppressed highly effectively (FPR = 0.024). However, the easy non-vascular negatives exhibit the highest false-positive rate (0.122) and the highest mean prediction score.

\section{Discussion}

The central clinical problem this work addresses is not detection sensitivity in the abstract,
but the overwhelming false-positive burden that prevents automated IA detection from reaching
the clinic. Current systems confuse aneurysmal sacs with normal vessel bifurcations because
both structures share nearly identical local intensity distributions in CTA, forcing radiologists
to manually review hundreds of spurious candidates per scan~\citep{bizjak2023systematic,hsu2025survey}.
Our results show that SECT substantially reduces this burden at the patch level by encoding
directional shape structure rather than intensity—giving the classifier access to information
that CNN-based systems fundamentally lack.

Unlike PI and PL, which achieve only moderate discrimination (AUC $\sim$0.68, Table~\ref{tab:main_results}) due to severe $H_2$ diagram collapse (85\% empty diagrams at $\varepsilon = 0.15$), SECT achieves robust classification (AUC 0.943 with RF, 0.931 with L1-SVC) by encoding the directional evolution of vascular geometry rather than individual feature lifespans. Crucially, SECT excels precisely where existing systems falter: it yields an AUC of 0.943 (78.4\% sensitivity at 95\% specificity) for sub-3~mm aneurysms, significantly outperforming typical CNN baselines. Furthermore, because SECT processes structural geometry rather than raw intensities, it demonstrates strong inter-scanner generalization, maintaining leave-one-scanner-out (LOGO) AUCs of 0.89--0.96 across four major manufacturers, albeit with minor residual scanner variance (Appendix~\ref{app:scanner_analysis}). Performance dips slightly for medium aneurysms (AUC 0.863), largely due to class imbalance and patch radius truncation for larger lesions.

SECT's discriminative power is also resilient to diverse negative cohorts, dropping only $\sim$0.015 AUC when evaluated on a primary clinical mixture. Notably, while it near-perfectly rejects challenging vascular bifurcations (FPR $= 0.003$), easy non-vascular negatives exhibit an anomalous FPR of 0.122, likely reflecting incidentally dome-like tissue cross-sections and RF miscalibration at the boundary. Although SECT's current inference time ($\sim$11~s/patch) is non-trivial for clinical pipelines, reducing to 64 directions and GPU parallelization offer clear optimization pathways. Future work will leverage these optimizations to integrate SECT into hybrid CNN architectures, enabling direct benchmarking of these robust topological priors against learned intensity representations.

\subsection{Limitations}

Evaluation is patch-level within a fixed candidate-generation setting; full end-to-end
pipeline validation remains future work. Size-strata imbalance limits conclusions for
medium and large aneurysms. Residual scanner-associated variation in the feature space
indicates partial rather than full scanner invariance. Computational cost at current
settings requires further engineering before real-time deployment.

\subsection{Conclusion}

We present a systematic evaluation of topological representations for intracranial aneurysm
detection. Persistence-based methods provide a meaningful but limited signal; SECT
delivers substantially higher discrimination by encoding directional geometric structure
that generalises across lesion size, scanner, and candidate distribution. The core clinical
takeaway is that shape-level representations can resolve the bifurcation ambiguity that
defeats intensity-based systems, and do so most reliably for the sub-3~mm lesions where
the clinical need is greatest. Future work will integrate SECT into full detection pipelines,
benchmark against CNN baselines, and pursue the computational optimisations needed for
clinical deployment.

\newpage
\bibliography{paper}

\newpage
\appendix
\section*{Appendix}

\section{Dataset Description}
\subsection{Scanner Distribution}
\label{app:scanner_dist}
The dataset exhibits realistic multi-scanner variability: Siemens: 377, GE: 333, Toshiba/Canon: 287, Phillips: 204. This diversity is critical for evaluating robustness to acquisition heterogeneity, a known limitation of CNN-based IA detection systems.

\subsection{Aneurysm Size Distribution}
\label{app:ia_size_dist}
The dataset is heavily skewed toward small aneurysms:  Small ($<3$ mm): 942 samples (78.4\%), Medium ($3$–$7$ mm): 112 samples (9.3\%), Large ($>7$ mm): 147 samples (12.3\%). This further strengthed out choice of the dataset as we can clearly perform size-stratified analysis; mainly focusing on small-aneurysms. See the method to find size estimate in \ref{app:size_verification}.

\section{Path Construction Details}
\label{app:patch_construction}
Patch extraction is performed on NIfTI volumes using a unified pipeline for positives and negatives. Inorder to reduce the scanner-dependent intensity variability; all the volumes first undergo an intensity pre-processing where: (i)Intensity clipping to $[0, 839]$ HU, (ii) Removal of air/background via thresholding ($<10$ HU) and (iii) Patch-level normalization using mean and standard deviation computed on non-air voxels.\\ \noindent
Patches are extracted using voxel spacing-aware cropping. All boundary conditions are handled via padding with minimum intensity values and the final patches are resized to a fixed resolution of 64x64x64. This ensures geometric consistency among patients with varying voxel spacings.

\begin{figure}[h!]
    \centering
    \includegraphics[width=0.95\linewidth]{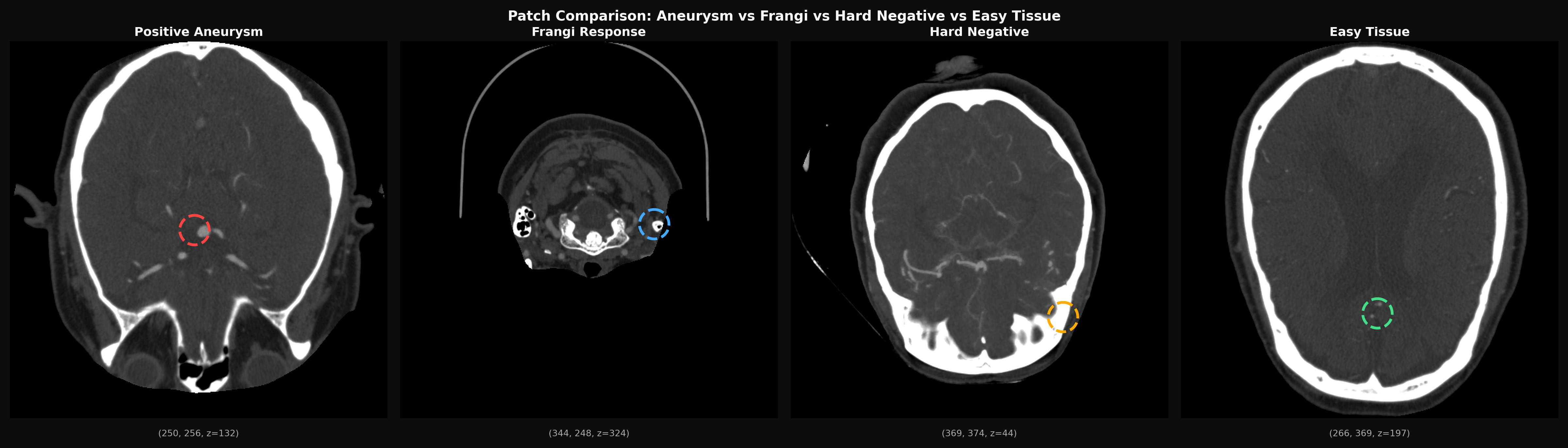}
    \caption{Representative examples of extracted CTA patches across classes. From left to right: (i) aneurysm-positive patch, (ii) Frangi-mined bifurcation candidate, (iii) hard vascular negative, and (iv) easy non-vascular tissue patch. Colored markers indicate patch centers used for extraction.
}
\label{fig:patch_construction}
\end{figure}

We construct a balanced dataset of positive and negative patches through a multi-stage procedure that enforces spatial, anatomical, and intensity constraints (see Fig.\ref{fig:patch_construction}).

\begin{algorithm}[h!]
\caption{Topology-Aware Extraction Sampling (TAXS) Strategy}
\label{alg:patch_construction}

\KwIn{CTA volume $V$, annotation CSV $\mathcal{A}$, radius $r = 15$ mm}
\KwOut{Set of labeled patches $\mathcal{P} = \mathcal{P}^+ \cup \mathcal{P}^-$}

\BlankLine
\textbf{Step 1: Positive Patch Extraction}\;

\ForEach{annotation entry $(x,y,\texttt{SOPInstanceUID}) \in \mathcal{A}$}{
    Reconstruct slice index $z$ using DICOM ordering via \texttt{ImagePositionPatient}\;
    Define center $c = (x,y,z)$\;
    Extract patch $P_c$ with physical radius $r = 15$ mm\;
    Add $P_c$ to $\mathcal{P}^+$\;
}

\BlankLine
\textbf{Step 2: Random Negative Sampling}\;

\ForEach{patient volume $V$}{
    Sample candidate locations $c'$ such that:\;
    \Indp
        $\|c' - c\|_2 > 40$ voxels for all positive centers $c$\;
        $c'$ lies within axial bounds $[0.15, 0.85]$ of volume depth\;
    \Indm

    \ForEach{candidate $c'$}{
        \eIf{intensity$(c') > 150$ HU and local maximum}{
            Add patch $P_{c'}$ to hard negatives $\mathcal{P}^{-}_{\text{hard}}$\;
        }{
            \If{$10 < \text{intensity}(c') < 150$}{
                Add patch $P_{c'}$ to easy negatives $\mathcal{P}^{-}_{\text{easy}}$\;
            }
        }
    }
}

\BlankLine
\textbf{Step 3: Frangi-Based Hard Negative Mining}\;

\ForEach{volume $V$}{
    Downsample $V$ to $25\%$ resolution\;
    Apply multi-scale Frangi filter with $\sigma \in \{1,2\}$\;
    Select top $0.2\%$ vesselness responses as candidate points\;
    Rescale coordinates to original resolution\;
    Extract corresponding patches and add to $\mathcal{P}^{-}_{\text{frangi}}$\;
}

\BlankLine
\textbf{Step 4: Sampling Balance}\;

\ForEach{patient}{
    Sample approximately $2 \times |\mathcal{P}^+|$ hard negatives\;
    Sample approximately $1 \times |\mathcal{P}^+|$ easy negatives\;
}

\BlankLine
\Return{$\mathcal{P}^+ \cup \mathcal{P}^{-}_{\text{hard}} \cup \mathcal{P}^{-}_{\text{easy}} \cup \mathcal{P}^{-}_{\text{frangi}}$}

\end{algorithm}
To ensure that topological descriptors are evaluated under clinically realistic failure modes, the designed patch construction pipeline (see Algorithm:\ref{alg:patch_construction}) explicitly targets vascular structures prone to false positives, particularly bifurcations.\\ 
For the frangi-mined negatives, instead of full Hessian-based vesselness computation at native resolution, we use a down sampled approximation to reduce computational cost. Empirical validation on a held-out subset confirms that this approximation preserves candidate quality without loss of relevant geometric information (see Appendix\ref{app:frangi_approx}).
\newpage
\section{Topological Vectorization}
\subsection{Persistence Images}
\label{app:pi}
Persistence diagrams (PDs) provide a compact summary of topological features but do not naturally reside in a vector space, making them difficult to integrate with standard machine learning models.Persistence Images (PI) \cite{adams2017persistence}, is a stable, finite-dimensional vector representation of PDs that enables the use of conventional learning algorithms.

In our pipeline, persistence images are computed from the $H_2$ persistence diagrams using:
\begin{itemize}
    \item Gaussian kernel with bandwidth $\sigma = 1.0$,
    \item Linear weighting function $w(b,p) = p$,
    \item Fixed grid resolution of $R \times R$ pixels over a normalized domain.
\end{itemize}

All infinite intervals are discarded prior to transformation.\\
\noindent Despite their practicality, persistence images introduce several limitations--- (i)Information loss due to discretization, (ii)Hyperparameter sensitivity and (iii)Loss of geometric structure. These limitations motivate the exploration of alternative representations that preserve more structural information while remaining stable and learnable.

\subsection{Persistence Landscapes}
\label{sec:pl}
Given a persistence diagram 
\begin{equation}
    PD = \{(b_i, d_i)\}_{i=1}^{n},
\end{equation}
each birth--death pair is associated with a \textbf{tent function}:
\begin{equation}
    f_{(b_i, d_i)}(t) =
    \begin{cases}
        0, & t \notin (b_i, d_i), \\
        t - b_i, & t \in (b_i, \tfrac{b_i + d_i}{2}], \\
        d_i - t, & t \in (\tfrac{b_i + d_i}{2}, d_i).
    \end{cases}
\end{equation}

Each function forms an isosceles triangle with peak height proportional to persistence and centered at the midpoint $(b_i + d_i)/2$. \\

\noindent The Persistence Landscapes (PL), introduced by Bubenik~\cite{bubenik2015statistical} is defined as a sequence of functions 
\begin{equation}
    \lambda_k : \mathbb{R} \rightarrow \mathbb{R}, \quad k = 1,2,\dots
\end{equation}
where
\begin{equation}
    \lambda_k(t) = \text{$k$-th largest value of } \{ f_{(b_i,d_i)}(t) \}_{i=1}^{n}.
\end{equation}

Equivalently, the persistence landscape can be viewed as a function
\begin{equation}
    \lambda : \mathbb{N} \times \mathbb{R} \rightarrow \mathbb{R}, \quad \lambda(k,t) = \lambda_k(t),
\end{equation}
forming a layered representation where --- $\lambda_1$ captures the most persistent feature at each scale, and higher-order $\lambda_k$ encode progressively weaker features.\\

% \paragraph{Functional Representation and Norms:}
% Persistence landscapes lie in $L^p$ spaces, typically $L^p(\mathbb{N} \times \mathbb{R})$, which are separable Banach spaces. This enables the use of standard vector-space operations such as norms, inner products, and averages:
% \begin{equation}
%     \|\lambda\|_p^p = \sum_{k=1}^{\infty} \|\lambda_k\|_p^p.
% \end{equation}

% This functional structure allows persistence landscapes to be treated as \textbf{random variables in Banach spaces}, making them amenable to statistical analysis. :contentReference[oaicite:2]{index=2}

\noindent In our pipeline, persistence landscapes are computed from $H_2$ persistence diagrams as follows:
\begin{itemize}
    \item All intervals are converted into triangular functions,
    \item Landscapes are truncated to the first $K$ levels,
    \item Functions are discretized over a fixed grid for numerical integration.
\end{itemize}

Despite their strong theoretical properties, PLs exhibit several limitations --- (i) Dimensional truncation, (ii) Loss of pointwise correspondence, and (iii) Smoothing of features. These limitations highlight the trade-off between statistical tractability and structural fidelity in functional representations of persistent homology.

\subsection{Smooth Euler Characteristic Transform (SECT) Formulation}
\label{app:sect}

The Smooth Euler Characteristic Transform (SECT), introduced by \citet{crawford2020predicting}, provides a principled geometric alternative to persistence-based summaries by mapping shapes into a collection of smooth functions. Each shape is represented as a set of functions that lie in a Hilbert space $\mathbb{L}^2$, equipped with a well-defined inner product.

\subsubsection*{Mathematical Formulation}
The SECT builds upon the Euler characteristic (EC), defined algebraically as the alternating sum of Betti numbers:
\begin{equation}
\chi(X) = \sum_{k=0}^{\infty} (-1)^k \beta_k,
\end{equation}
which compactly encodes the number of connected components, loops, and higher-dimensional voids. For a discretized complex $K$, this admits an equivalent combinatorial form as an alternating sum over cell counts across dimensions:
\begin{equation}
\chi(K) = \sum_{k=0}^{d} (-1)^k n_k,
\end{equation}
where $n_k$ denotes the number of $k$-dimensional cells. 

To capture multiscale directional structure, SECT considers a height function for a fixed direction $\nu \in S^{d-1}$:
\begin{equation}
    r_\nu(x) = x \cdot \nu,
\end{equation}
which induces a sublevel set filtration $K_\nu^x$. Tracking the EC across this filtration yields the Euler characteristic curve $\chi_\nu(x) = \chi(K_\nu^x)$. 

To ensure the representation is amenable to statistical learning, this piecewise constant curve must be smoothed. Following \citet{crawford2020predicting}, we first compute the mean-centered EC curve over the domain $[a_\nu, b_\nu]$:
\begin{equation}
Z_\nu^K(x) = \chi_\nu^K(x) - \bar{\chi}\nu^K.
\end{equation}
Integrating this centered curve produces a continuous, piecewise-linear functional representation known as the Smooth Euler Characteristic (SEC) curve:
\begin{equation}
F\nu^K(y) = \int_{-\infty}^{y} Z_\nu^K(x) dx.
\end{equation}
The full SECT representation is formally defined as the collection of these SEC curves over all sampled directions: $\nu \mapsto F_\nu^K \in \mathbb{L}^2$.

\subsubsection*{Properties and Limitations}
Compared to persistence landscapes and persistence diagrams, SECT offers several distinct representational advantages. It avoids the need for the arbitrary ordering or truncation of topological features, provides a globally consistent summary across spatial directions, and directly yields smooth functional representations without requiring auxiliary density transformations (such as Gaussian kerneling in Persistence Images). 

Furthermore, SECT carries strong theoretical guarantees, including injectivity and a well-defined Hilbert space structure that naturally permits norms, inner products, and functional regression. This mathematical foundation natively supports advanced functional data analysis (FDA) methods, including Gaussian process models, while ensuring the EC-based summaries capture rich, global shape information.

However, the application of SECT is bounded by certain computational and theoretical limitations. The expressivity of the transform is highly dependent on the density of the directional sampling over the sphere and exhibits sensitivity to the granularity of the filtration steps. Additionally, because the Euler characteristic acts as a global summary scalar at each filtration step, highly localized geometric details can occasionally be lost in the aggregation. These limitations reflect the inherent trade-off between statistical tractability and granular structural expressivity common to all functional summaries in topological data analysis.

% \subsection{Smooth Euler Characteristic Transform (SECT)}
% \label{app:sect}

% The Smooth Euler Characteristic Transform (SECT), introduced by \cite{crawford2020predicting}, provides a principled alternative by mapping shapes into a collection of smooth functions in a Hilbert space, enabling direct integration with statistical learning frameworks. Each shape is represented as a set of functions that lie in a Hilbert space $L^2$, equipped with a well-defined inner product.

% Compared to persistence landscapes and diagrams; SECT -- (i)avoids the need for ordering or truncation of features, (ii)provides globally consistent summary across directions, and (iii)directly yields smooth functional representations without any additional transformations. This makes SECT particularly suitable for integration into learning pipelines.

% It also follows the following theoretical guarantees -- (i)injectivity, (ii)Hilbert space structure allowing norms, functional regression along with inner products, (iii)supporting Gaussian process models and other FDA methods, and (iv) EC-based summaries capture global shape information.

% This makes SECT particularly suitable for integration into learning pipelines where inner products and kernels are required. However, it also has certain limitations like -- (i) dependence on directional sampling, (ii) filtration granularity sensitivity, and (iii) loss of local geometric details. These limitations reflect a trade-off between expressivity and statistical tractability, similar to other functional summaries in topological data analysis.

\begin{algorithm}[t]
\caption{SECT Feature Extraction for a CTA Patch}
\label{alg:sect}
\DontPrintSemicolon

\KwIn{Preprocessed patch $\tilde{P}_c \in \mathbb{R}^{64 \times 64 \times 64}$, vascular threshold $\tau_v = 0.40$, directions $D=128$, steps $T=25$}
\KwOut{SECT feature vector $\xi \in \mathbb{R}^{3200}$}

\textbf{1. Normalize patch:} Apply preprocessing (Sec.~3.2) to obtain $f : \tilde{P}_c \rightarrow [0,1]$\;

\textbf{2. Vascular mask:} $M = \{x \in \tilde{P}_c : f(x) \geq \tau_v\}$ \tcp*{$\tau_v \neq$ PH threshold $\varepsilon$}

\textbf{3. Sample directions:} Generate $\{v_i\}_{i=0}^{D-1} \subset S^2$ via Fibonacci sphere:
\[
y_i = 1 - \frac{2i}{D-1}, \;
r_i = \sqrt{1-y_i^2}, \;
\theta_i = \phi i, \;
\phi = \pi(3-\sqrt{5})
\]
\[
v_i = (r_i\cos\theta_i,\; y_i,\; r_i\sin\theta_i)
\]

\For{$k \leftarrow 1$ \KwTo $D$}{
    \textbf{4. Height function:} $h_{v_k}(x) = v_k \cdot x$, normalized to $[-1,1]$\;

    \textbf{5. Sublevel sweep:} $\{t_j\}_{j=1}^T = \texttt{linspace}(-1,1,T)$\;
    
    \For{$j \leftarrow 1$ \KwTo $T$}{
        $A_{v_k}(t_j) = M \cap \{x : h_{v_k}(x) \leq t_j\}$\;
        $\chi_{k,j} = \chi(A_{v_k}(t_j))$ using 26-connectivity\;
    }

    \textbf{6. Smooth:} $\tilde{\chi}_{k,:} = \texttt{GaussianSmooth}(\chi_{k,:}, \sigma=1.0)$\;
}

\textbf{7. Output:} $\Xi \in \mathbb{R}^{D \times T}$, $\xi = \texttt{vec}(\Xi) \in \mathbb{R}^{3200}$, $\ell_2$-normalized\;

\Return{$\xi$}
\end{algorithm}
\section{Hyper-parameter Selection}
\label{app:hyperparams}
For hyper-parameter selection, and to test the stability of extracted features, from the full dataset we constructed a curated subset consisting of – 154 positive patient patches (each containing at least one annotated aneurysm), 394 Frangi-mined hard negatives, 221 random hard negatives, 145 easy tissue negatives.
\subsection{Intensity Normalization and Clipping}
\label{app:eda_clipping}

To determine an appropriate intensity normalization range, we perform an exploratory analysis of voxel intensities across a representative subset of the dataset. For each volume, we compute summary statistics including minimum, maximum, mean, and upper percentiles of Hounsfield Units (HU).

We observe that while raw intensities vary across scans, the majority of informative signal is concentrated within a bounded range, with extreme values corresponding to noise or acquisition artifacts. In particular, the $99^{\text{th}}$ percentile (P99) provides a stable upper bound across samples.

Based on this analysis, we adopt a clipping range of $[0, 839]$ HU, which retains the relevant vascular and soft-tissue contrast while suppressing extreme outliers. This choice is consistent with the empirically observed intensity distribution and is used uniformly across all experiments.

\subsection{Validation of Fast Frangi Approximation}
\label{app:frangi_approx}

To enable large-scale hard negative mining, we evaluate a computationally efficient approximation of the Frangi vesselness filter. The reference implementation applies the filter at $0.5\times$ resolution with $\sigma \in \{1,2,3\}$, while the proposed approximation operates at $0.25\times$ resolution with $\sigma \in \{1,2\}$, resulting in a $\sim 4\times$ reduction in processing time.

Since the two methods produce spatially independent candidate patches, we assess equivalence by analyzing the structural properties of the extracted patches themselves. For each method, a subset of patches is evaluated using in-patch vesselness and intensity-based metrics, including the $95^{\text{th}}$ percentile Frangi response, mean vesselness, vessel occupancy (fraction of voxels $>150$ HU), mean HU, and intensity standard deviation.

Statistical comparison using a two-sample Kolmogorov--Smirnov test shows no significant difference between the two distributions across all metrics ($p > 0.05$). The primary hardness measure, in-patch Frangi P95, yields $p = 0.068$ with negligible Wasserstein distance, indicating near-identical structural content.

\subsection{Persistence Diagram Preprocessing Selection}
\label{app:pd_grid}

To determine appropriate preprocessing for persistence diagram (PD) computation, we perform a grid search over key parameters controlling intensity normalization and smoothing. The evaluated grid includes Gaussian smoothing ($\sigma \in \{0.0, 0.5, 1.0\}$), intensity clipping ($\text{max} \in \{800, 1000, 1200\}$), and normalization method (z-score vs.\ min-max).

Each parameter combination is evaluated on a balanced subset of positive, Frangi, hard, and easy patches. Performance is assessed using a topology-based separability ratio (TSR), defined as the ratio of inter-class distance (positive vs.\ Frangi) to intra-class distance (positive vs.\ positive), computed using bottleneck distance.

The optimal configuration corresponds to Gaussian smoothing $\sigma = 0.5$, intensity clipping at $800$ HU, and min-max normalization, which maximizes the TSR while maintaining stable intra-class structure. These parameters are used for all subsequent persistence-based analyses.

\subsection{SECT Parameter Sensitivity and Robustness}
\label{app:sect_params}
SECT hyperparameters are selected via a grid search over the number of projection directions ($n_{\text{dirs}} \in \{32, 64, 128, 256\}$), filtration resolution ($t_{\text{steps}} \in \{25, 50, 100\}$), and smoothing strategy. The optimal configuration ($128$ directions, $25$ steps, Gaussian smoothing) provides the best trade-off between performance and computational cost.

\begin{figure}[h]
\centering
\includegraphics[width=0.75\linewidth]{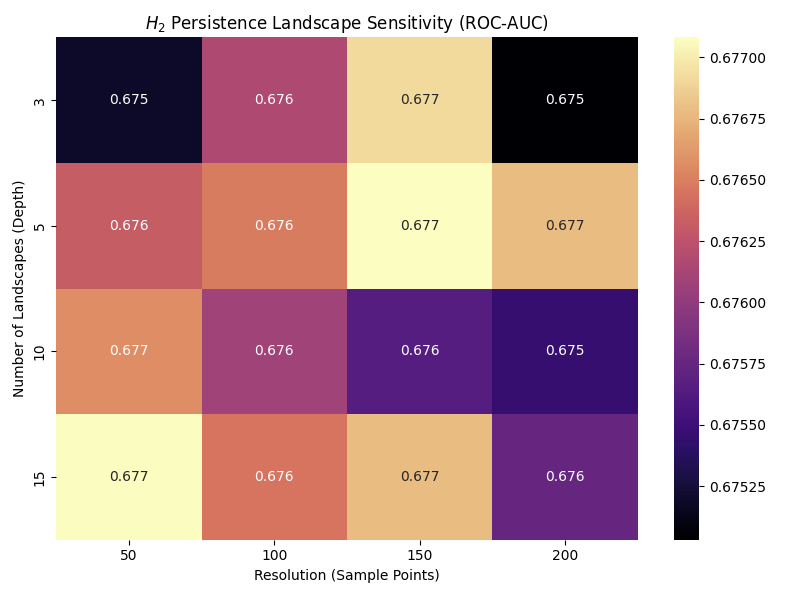}
\caption{
Sensitivity of persistence landscape features to resolution and number of landscapes. 
Performance remains stable across all configurations, indicating low dependence on discretization choices.
}
\label{fig:pl_sensitivity}
\end{figure}

\begin{figure}[h]
\centering
\includegraphics[width=0.75\linewidth]{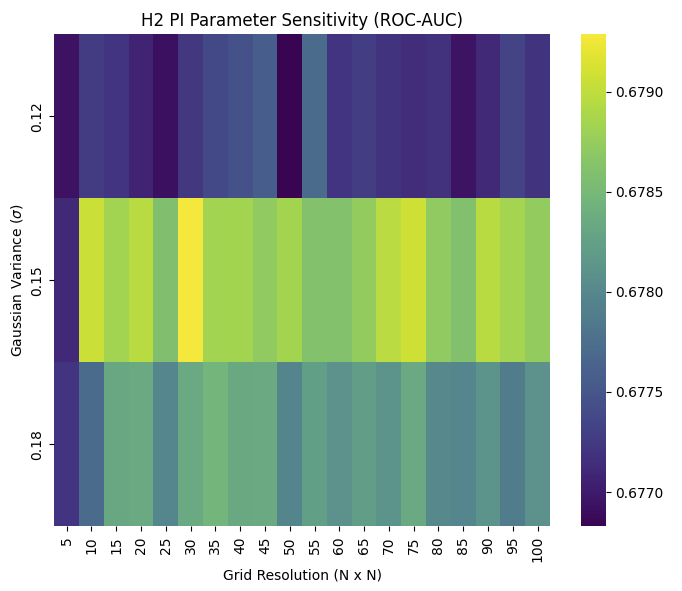}
\caption{
Sensitivity of persistence image features to grid resolution and Gaussian variance. 
AUC varies minimally across configurations, demonstrating robustness to preprocessing parameters.
}
\label{fig:pi_sensitivity}
\end{figure}

To assess robustness, we analyze performance across a broad range of parameter settings. As shown in Fig.~\ref{fig:pl_sensitivity}, SECT-derived representations exhibit minimal variation across resolution and landscape depth, with AUC remaining effectively constant ($\approx 0.675$–$0.677$). Similarly, Fig.~\ref{fig:pi_sensitivity} shows that persistence image performance remains stable across grid resolutions and Gaussian variances, with AUC variation below $0.01$.


We additionally evaluate sensitivity to the vascular mask threshold $\tau_v \in \{0.30, 0.40, 0.50, 0.60\}$. Across all settings, performance remains unchanged (AUC $=0.937$), indicating that SECT features are insensitive to moderate variations in thresholding.

\paragraph{Summary.}
These results demonstrate that SECT operates in a stable regime, where performance is largely invariant to reasonable choices of hyperparameters. This robustness suggests that the method does not rely on fine-tuning and generalizes consistently across parameter configurations.

\section{Ablation Studies}
\subsection{Topological Separability via Persistence Diagrams}
\label{app:topo_seperability}
% maybe we can move this experiment to appendix and also then keep the parameter selection associated with it only.

\begin{figure}[htbp]
\centering
\begin{minipage}[c]{0.42\textwidth}
\centering
\small
\begin{tabular}{l c}
\toprule
\textbf{Comparison} & \textbf{Mean $d_B$ ($\downarrow$)} \\
\midrule
Pos vs. Pos (Intra) & 0.0418 $\pm$ 0.0683 \\
Pos vs. Frangi & 0.0748 $\pm$ 0.0816 \\
Pos vs. Hard & 0.1323 $\pm$ 0.0888 \\
\bottomrule
\end{tabular}
\end{minipage}
\hfill
\begin{minipage}[c]{0.54\textwidth}
\centering
\includegraphics[width=\linewidth]{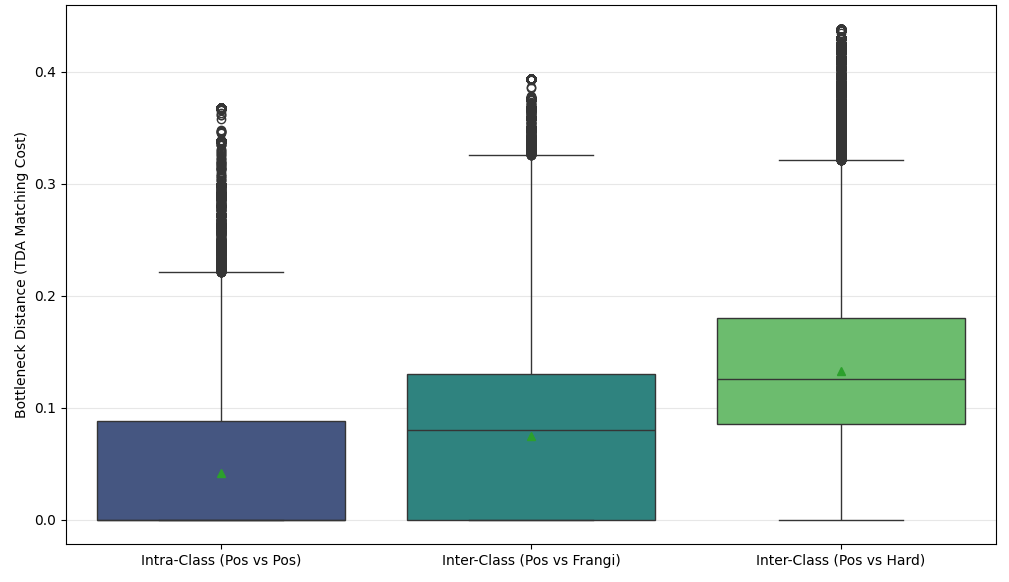}
\end{minipage}
\caption{\textbf{Empirical Topological Separability via $H_2$ Bottleneck Distance.} The table (left) reports mean distances $\pm$ standard deviation, while the box plot (right) illustrates the full distributions across comparison groups (green triangles denote distribution means; horizontal lines denote medians). Aneurysm patches exhibit the highest topological similarity to each other (lowest $d_B$), while Frangi-mined bifurcations occupy an intermediate regime, mathematically validating their role as the primary geometric confounder in CNN-based detection.}
\label{fig:topo_sep}
\end{figure}

Before evaluating downstream classification, we first investigate whether persistent homology natively captures a meaningful geometric signal capable of distinguishing aneurysm patches from vascular structures. For each patch, we compute $H_2$ persistence diagrams using a cubical complex under a superlevel filtration. We evaluate topological similarity using the pairwise bottleneck distance, $d_B(\cdot, \cdot)$, across three distinct comparison groups: (i) intra-class (Aneurysm vs. Aneurysm), (ii) inter-class hard-negatives (Aneurysm vs. Frangi-mined bifurcations), and (iii) non-aneurysmal vascular inter-class (Aneurysm vs. Random Hard Negatives). The table in Figure~\ref{fig:topo_sep} reports the mean bottleneck distances across $N=1201$ samples per class at persistence threshold $\varepsilon=0.15$.

We observe a consistent distance hierarchy:
\[
d_B(\text{Pos}, \text{Pos}) < d_B(\text{Pos}, \text{Frangi}) < d_B(\text{Pos}, \text{Hard}),
\]

This hierarchy, which remains stable across persistence thresholds $\varepsilon \in [0.10, 0.20]$, confirms that aneurysm patches are topologically more similar to each other than to vascular structures. Furthermore, Frangi-mined bifurcations occupy an intermediate topological class.

These results provide empirical evidence that persistent homology captures a non-trivial geometric signal aligned with anatomical structure. Importantly, this signal does not correspond to persistently enclosed cavities, but rather reflects the differences in local geometric structure under the filtration. While global separability is observed at the pairwise distance level, the underlying $H_2$ topological signal is inherently sparse and highly sensitive to noise thresholding. A detailed diagnostic audit of threshold selection and $H_2$ feature degeneracy is provided in Appendix \ref{app:ph_diagnostic}.

\subsection{Details about Topological Separability and Degeneracy}
\label{app:ph_diagnostic}
While we try to answer the question (i)do aneurysm-positive patches induce a coherent geometry in PD space, i.e., are positives closer to each other than to hard negatives?; it is also necessary to answer (ii)how often do persistence channels collapse to empty diagrams under denoising thresholds, and how is this affected by homology dimension and negative type?

We evaluate: (i)aneurysm-positive (\texttt{positive}), Frangi-based candidate negatives (\texttt{fast\_frangi}), and a negative baseline (\texttt{negative\_hard}).

To quantify class geometry directly in diagram space, we use the bottleneck distance $d_B(\cdot,\cdot)$ between $H_2$ persistence diagrams.

For each class and threshold, we report the \emph{degeneracy rate} in each homology dimension,
defined as the fraction of patches whose thresholded diagram is empty:
$
\delta_k(\varepsilon) = \Pr\big(|D_k| = 0\big).
$
We additionally report the mean surviving feature count $\mathbb{E}[|D_k|]$.
This audit is essential, since high degeneracy implies that any PD-vectorisation (including PI/PL) may be dominated by collapse patterns rather than rich persistence geometry. We also report the Topological Separability Ratio (TSR),
\[
\mathrm{TSR}(\varepsilon) =
\frac{\mathbb{E}[d_B(\mathrm{Pos},\mathrm{Frangi})]}
     {\mathbb{E}[d_B(\mathrm{Pos},\mathrm{Pos})]},
\]
which remains $>1$ for both thresholds ($\mathrm{TSR}_{0.15}\approx 1.79$, $\mathrm{TSR}_{0.20}\approx 1.93$), confirming that inter-class separation is not merely intra-class noise.\\
\noindent For positives, $H_2$ degeneracy rises from $77.10\%$ at $\varepsilon=0.10$ to $85.01\%$ at $\varepsilon=0.15$ and $90.34\%$ at $\varepsilon=0.20$, while the mean surviving $H_2$ counts drop from $2.85$ to $0.58$ to $0.23$ respectively.
In contrast, negative classes retain more $H_2$ structure under the same thresholds.
Notably, $H_1$ degeneracy is consistently lower than $H_2$ degeneracy in positives, suggesting that if persistence is to be used, loop-like structure ($H_1$) may carry more stable signal than volumetric voids ($H_2$) under thresholding.

\begin{table}[t]
\centering
\caption{Empirical $H_2$ bottleneck separability across persistence thresholds $\varepsilon$.}
\label{tab:h2_bottleneck_separability}
\begin{tabular}{lcccc}
\toprule
$\varepsilon$ &
$\mathbb{E}[d_B(\mathrm{Pos},\mathrm{Pos})]$ &
$\mathbb{E}[d_B(\mathrm{Pos},\mathrm{Frangi})]$ &
$\mathbb{E}[d_B(\mathrm{Pos},\mathrm{Hard})]$ &
$\mathrm{TSR}(\varepsilon)$ \\
\midrule
0.15 & 0.0418 & 0.0748 & 0.1323 & 1.79 \\
0.20 & 0.0311 & 0.0600 & 0.1168 & 1.93 \\
\bottomrule
\end{tabular}
\end{table}
\begin{figure}
    \centering
    \includegraphics[width=0.95\linewidth]{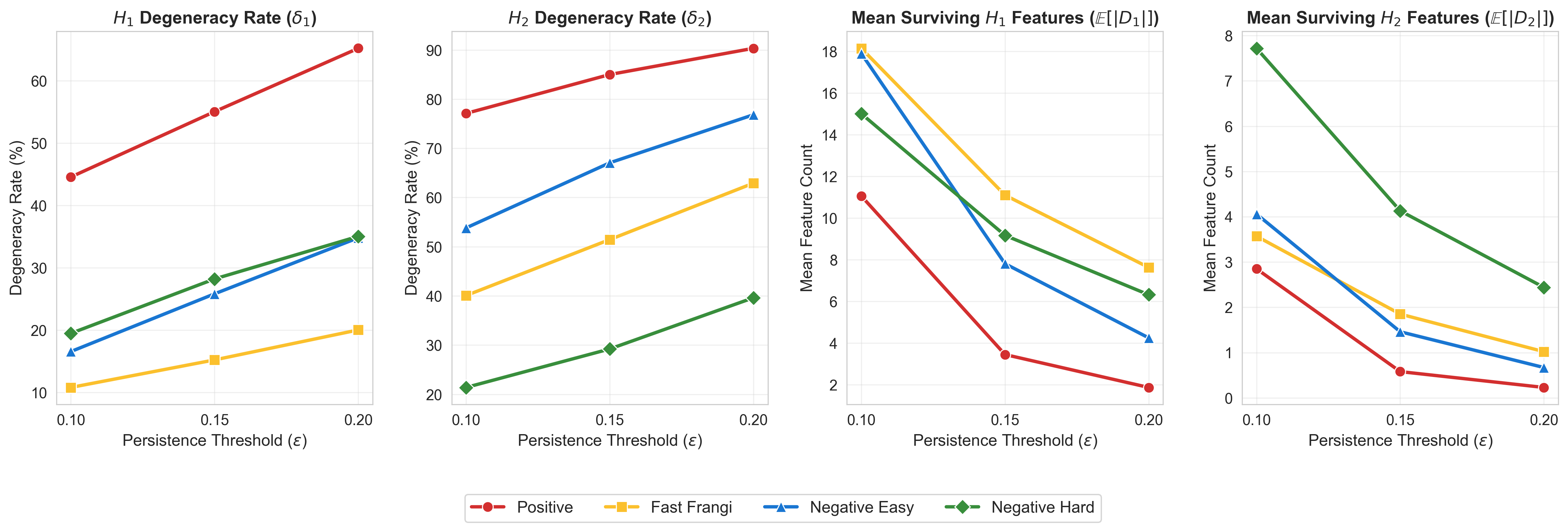}
    \caption{Topological degeneracy and feature survival across persistence thresholds ($\varepsilon$):From left to right, the panels illustrate the degeneracy rates—the percentage of patches yielding empty persistence diagrams—for 1-dimensional ($H_1$) and 2-dimensional ($H_2$) topological features, followed by the mean number of surviving features per diagram for $H_1$ and $H_2$, respectively. Notably, true Positive aneurysm patches exhibit significantly higher degeneracy rates and lower surviving feature counts compared to Frangi, Hard, and Easy negative patches. This divergence underscores the morphological simplicity of true saccular aneurysms, which consist of isolated, prominent structures, whereas false positive mimics contain complex, persistent structural noise.}
    \label{fig:topological_degeneracy}
\end{figure}
% \begin{table}[t]
% \centering
% \caption{Degeneracy rates and mean surviving feature counts at thresholds $\varepsilon\in\{0.10,0.15,0.20\}$.
% We report $\delta_k$ as the percentage of empty diagrams after thresholding, and $\mathbb{E}[|D_k|]$ as the mean surviving count.}
% \label{tab:degeneracy_audit}
% \begin{tabular}{llcccc}
% \toprule
% $\varepsilon$ & Class &
% $\delta_1(\%)$ & $\delta_2(\%)$ &
% $\mathbb{E}[|D_1|]$ & $\mathbb{E}[|D_2|]$ \\
% \midrule
% 0.10 & Positive      & 44.55 & 77.10 & 11.06 & 2.85 \\
% 0.10 & Fast Frangi   & 10.79 & 40.07 & 18.15 & 3.57 \\
% 0.10 & Negative Easy & 16.55 & 53.82 & 17.90 & 4.06 \\
% 0.10 & Negative Hard & 19.47 & 21.38 & 15.01 & 7.72 \\
% \midrule
% 0.15 & Positive      & 55.04 & 85.01 & 3.45  & 0.58 \\
% 0.15 & Fast Frangi   & 15.21 & 51.45 & 11.10 & 1.85 \\
% 0.15 & Negative Easy & 25.81 & 67.06 & 7.81  & 1.46 \\
% 0.15 & Negative Hard & 28.20 & 29.21 & 9.17  & 4.13 \\
% \midrule
% 0.20 & Positive      & 65.20 & 90.34 & 1.87  & 0.23 \\
% 0.20 & Fast Frangi   & 20.02 & 62.92 & 7.63  & 1.02 \\
% 0.20 & Negative Easy & 34.80 & 76.83 & 4.25  & 0.67 \\
% 0.20 & Negative Hard & 35.03 & 39.62 & 6.32  & 2.44 \\
% \bottomrule
% \end{tabular}
% \end{table}

% \begin{figure}[t]
%     \centering
%     \includegraphics[width=0.95\linewidth]{images/topological_separability_results.png}
%     \caption{Distribution of $H_2$ bottleneck distances for intra-class positives and inter-class comparisons.
%     We report these as descriptive distributions due to pairwise dependence.
%     }
%     \label{fig:h2_bottleneck_boxplot}
% \end{figure}

\subsection{Patch-Level Aneurysm Size Verification}
\label{app:size_verification}
Aneurysm size annotations are derived using a dedicated patch-driven pipeline that estimates the physical size of each aneurysm directly from imaging metadata. Unlike prior approaches that aggregate size at the patient or series level, this pipeline produces one size estimate per patch, ensuring correct handling of multi-aneurysm cases. As a result, performance differences observed across size strata can be attributed to representation quality rather than annotation artifacts.

Each patch is assigned to a size category using clinically standard thresholds-- (i)\textbf{Small:}$<3$ mm, (ii)\textbf{Medium:}$3-7$mm and (iii)\textbf{Large:}$>7$ mm. These thresholds align with prior clinical literature and reflect known differences in detection difficulty across aneurysm sizes.

Each positive patch is identified by a unique \texttt{patch\_id}, derived from its filename of the form:\texttt{<SeriesInstanceUID>\_positive\_<x>\_<y>\_<z>}. The spatial coordinates $(x,y,z)$ are used to match each patch to its corresponding aneurysm annotation in the source dataset. Matching is performed in two stages:
\begin{itemize}
    \item \textbf{Series-level filtering:} Candidate annotations are first restricted to the same \texttt{SeriesInstanceUID}.
    \item \textbf{Spatial association:} The patch location is matched to the closest annotation centroid in the $(x,y)$ plane, optionally using aneurysm labels when available, or otherwise via proximity within a fixed tolerance.
\end{itemize}

This procedure ensures that each patch is associated with the correct aneurysm, even in the presence of multiple lesions within the same scan. :contentReference[oaicite:0]{index=0}

Aneurysm size is estimated in millimeters using a 3D bounding-box proxy derived from both in-plane and through-plane measurements:

\begin{itemize}
    \item \textbf{XY diameter:} Computed as the spatial extent of annotation coordinates in the image plane, scaled by voxel spacing obtained from NIfTI headers.
    \item \textbf{Z height:} Computed from slice positions using DICOM \texttt{ImagePositionPatient} metadata. When full slice coverage is unavailable, fallback strategies based on slice thickness are used.
\end{itemize}

To provide a robust upper-bound estimate of aneurysm extent in 3D; the final size estimate (dome diameter proxy) is defined as:\[
\text{Dome}_{\text{mm}} = \max(\text{XY diameter}, \text{Z height}),
\]

To ensure robustness across heterogeneous imaging data, the pipeline incorporates multiple fallback strategies --- (i) If NIfTI spacing is unavailable, pixel-based XY measurements are used, (ii)If DICOM slice positions are incomplete, Z-extent is approximated using slice count and thickness, and (iii) If slice thickness is missing, a default physical thickness is used.

Patches for which no valid annotation can be matched are excluded from stratified.

To assess statistical differences across strata, we perform a Kruskal–Wallis test on predicted scores, which indicates a significant difference across size groups ($H=37.23$, $p<0.001$). Pairwise DeLong tests show that performance for small aneurysms is significantly higher than medium ($\Delta$AUC $=0.0763$, $p=0.0009$), while differences between medium and large are not statistically significant.

\begin{figure}
    \centering
    \includegraphics[width=0.95\linewidth]{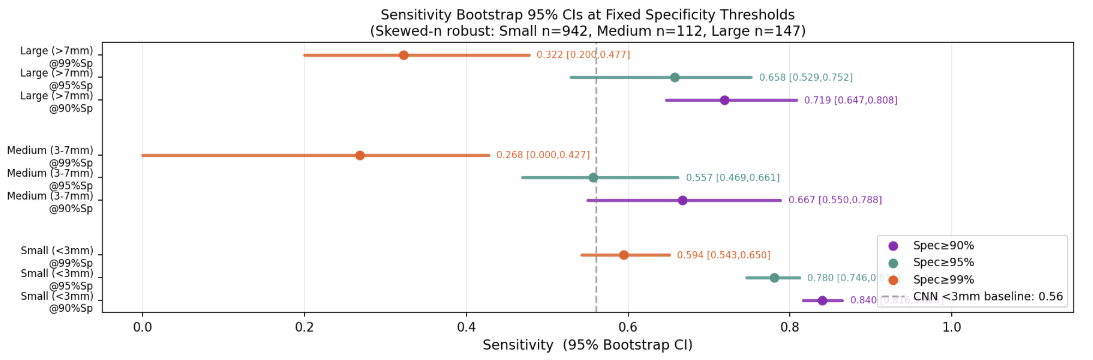}
    \caption{Size-stratified SECT performance with 95\% bootstrap confidence intervals. 
Confidence intervals widen for medium and large strata due to limited sample size, 
while small aneurysms exhibit stable estimates despite class imbalance.}
    \label{fig:size_ci}
\end{figure}

Sensitivity at fixed specificity further reflects the effect of sample size (see Fig.\ref{fig:size_ci} ) -- confidence intervals remain tight for small aneurysms but widen substantially for medium and large strata, consistent with reduced statistical stability under limited sample sizes.. Diagnostic logging confirms near-complete correspondence between patch files and size entries.

\subsection{Methodology and Domain Definitions for Empirical Lipschitz Analysis}
\label{app:lipschitz_methodology}

To evaluate the robustness of our topological representations against scanner noise, we conducted an empirical Lipschitz stability analysis. We applied uniform physical perturbations to the source data by injecting Gaussian noise $\mathcal{N}(0, \sigma^2)$ directly into the normalized 3D CTA scalar fields, sweeping $\sigma \in \{0.01, 0.05, 0.10, 0.15, 0.20\}$.

It is critical to note that the empirical Lipschitz constant $L$ is calculated across different input metric spaces depending on the representation. For Persistence Images (PI) and Persistence Landscapes (PL), stability is traditionally bounded relative to the intermediate Persistence Diagram (PD). Let $W_p(D_1, D_2)$ denote the $p$-Wasserstein distance between diagrams. The empirical Lipschitz constant for PI and PL is calculated as:
$$ L_{\Phi} \approx \max \frac{||\Phi(D_{base}) - \Phi(D_{noisy})||_2}{W_p(D_{base}, D_{noisy})} \quad \text{for } \Phi \in \{\text{PI}, \text{PL}\} $$

Conversely, the Smooth Euler Characteristic Transform (SECT) operates directly on the scalar field and does not compute a persistence diagram. Therefore, its stability is measured end-to-end against the intensity perturbation domain:
$$ L_{SECT} \approx \max \frac{||\text{SECT}(X_{base}) - \text{SECT}(X_{noisy})||_2}{\sigma} $$

Because the denominators occupy different metric spaces (Wasserstein distance versus image intensity variation), the raw numerical $L$ values cannot be compared directly. However, analyzing them within their respective domains reveals important behavioral differences: PI exhibits extreme sensitivity (high amplification) to minor topological shifts $L \approx 430.5$, whereas SECT demonstrates bounded, moderate variation $L \approx 40.3$ directly with respect to physical image noise.

\subsection{Synthetic Validation of SECT under Controlled Geometric Perturbations}
\label{app:phantom_validation}

To validate that SECT captures intrinsic geometric differences between aneurysm-like and bifurcation-like structures, we conduct controlled synthetic experiments using parameterized 3D phantoms.

We generate two classes of volumetric phantoms: (i) \textbf{saccular structures}, modeled as a spherical outpouching attached to a parent vessel, and (ii) \textbf{bifurcation structures}, modeled as a branching tubular configuration. All phantoms are embedded in a fixed $32^3$ grid, with geometry controlled via scale, orientation, and noise.

We systematically vary three factors--- (i)\textbf{Noise:} $\sigma \in \{0.0, 0.15, 0.30\}$ (clean, moderate, heavy), (ii)\textbf{Scale:} $s \in \{0.6, 1.0, 1.5\}$ (small, baseline, large), and (iii)\textbf{Orientation:} fixed, limited ($\pm30^\circ$), and random ($[0^\circ, 360^\circ)$). This results in $3 \times 3 \times 3 = 27$ conditions. For each condition, we generate $n=60$ samples per class and compute SECT features using the same configuration as the main experiments. Classification is performed using a Random Forest with 5-fold cross-validation, and performance is reported as mean AUC with bootstrap confidence intervals.

SECT consistently separates saccular and bifurcation phantoms across most conditions. Under clean and moderate noise, performance is near-perfect (AUC $\approx 0.99$ across all scales and orientations). Under heavy noise, performance degrades but remains above chance in all cases, with mean AUC $0.798 \pm 0.117$.

Across all 27 conditions, the overall mean AUC is $0.9295 \pm 0.1152$. Performance decreases with increasing noise and random orientation, particularly for small-scale structures, but remains strongly discriminative in most regimes.

\begin{figure}
    \centering
    \includegraphics[width=0.95\linewidth]{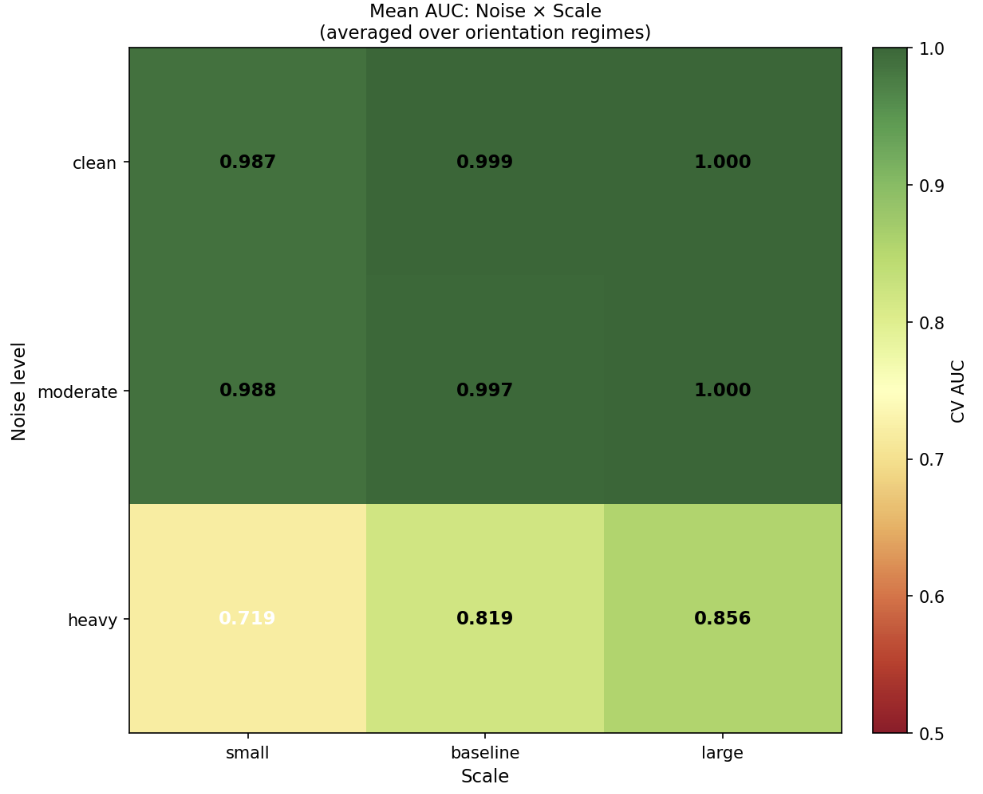}
    \caption{Synthetic validation of SECT under controlled perturbations. Heatmap shows mean AUC for distinguishing saccular and bifurcation phantoms across noise and scale conditions (averaged over orientations). SECT achieves near-perfect separation under clean and moderate noise, with degradation under heavy noise, particularly for small-scale structures.}
    \label{fig:sect_geometry}
\end{figure}
These results demonstrate that SECT captures stable geometric signatures distinguishing saccular and bifurcation structures, and that this signal persists under realistic perturbations in scale, orientation, and noise. Performance degradation under extreme noise conditions is expected, reflecting loss of topological structure in the underlying signal rather than failure of the representation.

\subsection{Inter-Scanner Metadata and Feature-Space Analysis}
\label{app:scanner_analysis}
For each unique imaging series, a single DICOM file is accessed to extract acquisition metadata, including manufacturer and model information. Manufacturer names are normalized into canonical groups (Siemens, GE, Philips, Toshiba/Canon) to account for variations in raw DICOM strings.

Each positive and negative patch, inherits the scanner label of its source series, resulting in a consistent patch-level assignment without requiring full-series reconstruction. This strategy ensures efficient metadata extraction while preserving correct scanner attribution across all samples.

The dataset exhibits substantial diversity across scanner manufacturer (see Appendix\ref{app:scanner_dist}) which reflects a multi-institutional acquisition setting and provides a suitable basis for evaluating scanner robustness.

\subsubsection{Feature-Space Analysis.}
\begin{figure}
    \centering
    \includegraphics[width=0.65\linewidth]{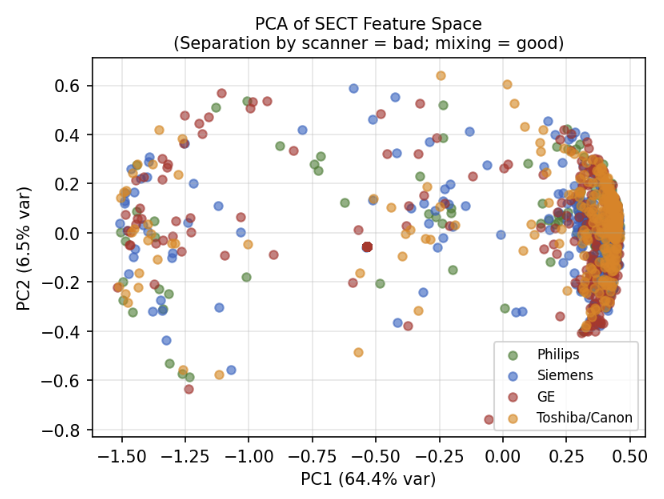}
    \caption{The scatter plot visualizes the first two principal components (accounting for 64.4\% and 6.5\% of the variance, respectively) for positive aneurysm patches sourced from a multi-institutional dataset.}
    \label{fig:pca_scanner_robustness}
\end{figure}

To assess whether SECT embeddings encode scanner-specific structure, we analyze the feature space using Principal Component Analysis (PCA). High-dimensional SECT vectors are projected onto a lower-dimensional subspace, and scanner labels are examined for clustering behavior. If SECT representations are scanner-agnostic, samples from different scanners should exhibit substantial overlap in the projected feature space rather than forming distinct clusters. Fig.\ref{fig:pca_scanner_robustness} shows a heavy intertwining of distributions across Siemens, GE, Philips, and Toshiba/Canon scanners indicating strong feature-level mixing; although mild scanner-associated shifts are observable in lower-dimensional projections.
\subsubsection{ANOVA-Based Scanner Effect Quantification.}
To quantify scanner-associated variation, we perform one-way ANOVA across scanner groups on the leading principal components. Specifically, we evaluate whether the distribution of feature values differs significantly across scanners for each principal component. The analysis yields the following results:
\begin{itemize}
    \item Mean F-statistic (first 10 PCs): $5.42$
    \item Number of significant PCs ($p < 0.05$): $7/10$
\end{itemize}

The observed F-statistics indicate the presence of moderate scanner-associated variation in the SECT feature space. However, qualitative inspection of PCA projections shows substantial mixing of samples across scanner groups, with no clear separation by manufacturer. This suggests that while scanner identity contributes to feature variance, it does not dominate the representation. In particular, scanner-specific variation appears secondary to the underlying class-discriminative structure captured by SECT.

\subsection{Detailed Operating-Point and False Positive Analysis}
\label{app:fp_analysis}

To assess the clinical viability of the topological representations, we conducted a granular operating-point analysis focusing on the high-sensitivity regime strictly required for intracranial aneurysm screening. Using out-of-fold predictions from a 5-fold cross-validation setup, we evaluated the False Positive Rate (FPR) at a fixed 90\% sensitivity threshold. Additionally, we computed a bifurcation-specific FPR (utilizing the Youden-optimal threshold) to isolate the model's performance on the most challenging morphological mimics: healthy vessel bifurcations.

\begin{figure*}[htbp]
    \centering
    \includegraphics[width=\textwidth]{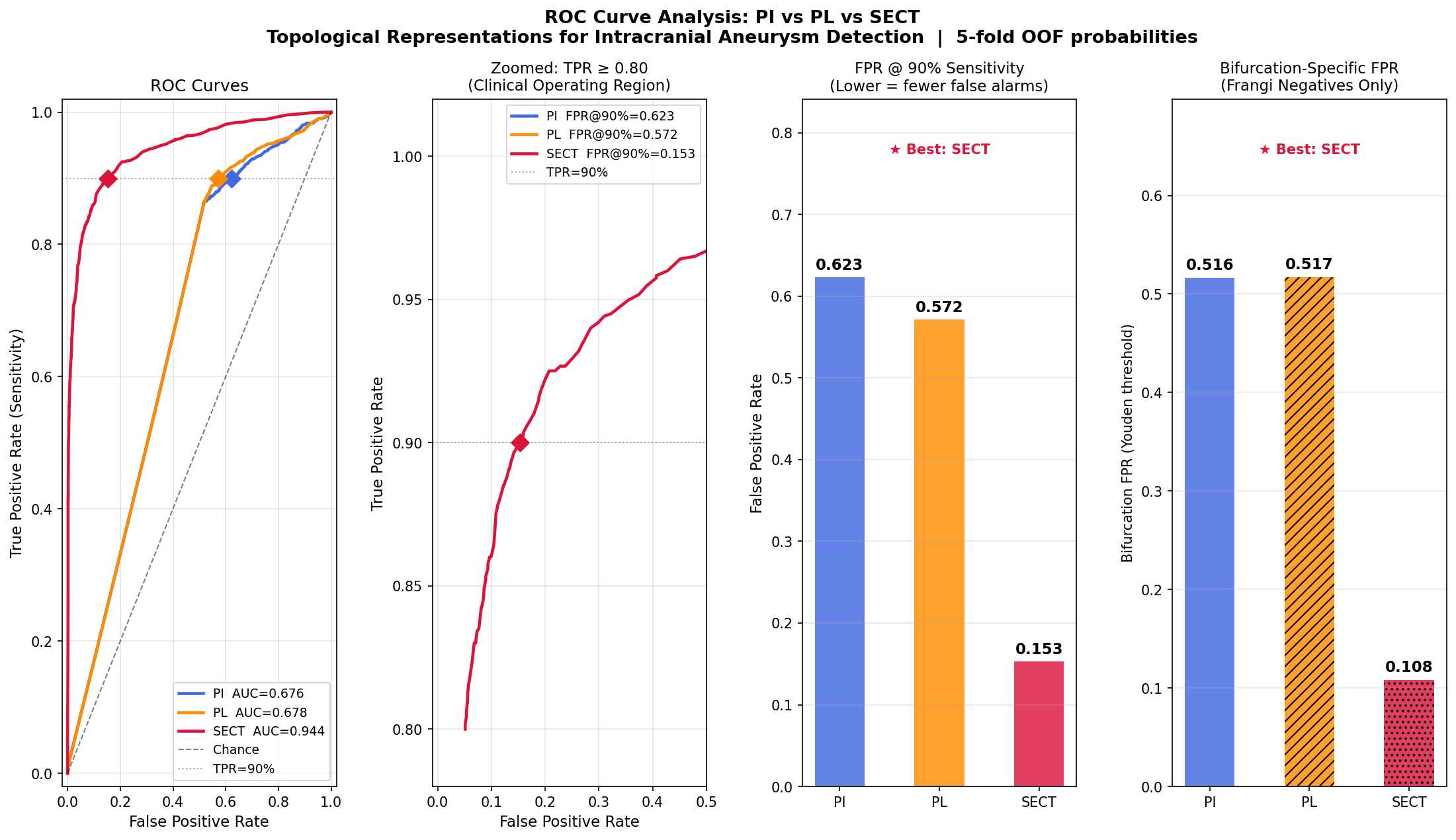}
    \caption{\textbf{Detailed Operating-Point and False Positive Analysis.} (Left to Right) \textbf{Panel 1:} Standard ROC curves across the three topological representations. \textbf{Panel 2:} Zoomed ROC focusing on the clinically critical high-sensitivity regime (TPR $\ge$ 0.80). \textbf{Panel 3:} False Positive Rate (FPR) forced at a strict 90\% sensitivity. SECT reduces the false alarm rate to 15.3\%, avoiding the specificity collapse seen in PI and PL. \textbf{Panel 4:} Bifurcation-specific FPR measured exclusively on Frangi-mined hard negatives, confirming SECT's superior capacity to distinguish true saccular geometry from confounding vascular branches.}
    \label{fig:roc2}
\end{figure*}

As illustrated in the zoomed ROC analysis (Figure \ref{fig:roc2}, Panel 2), SECT maintains robust discriminative power even in the aggressive True Positive Rate (TPR) $\ge$ 80\% regime. At a fixed 90\% sensitivity, SECT achieves an FPR of 0.153 (Panel 3). In stark contrast, the direction-agnostic baselines suffer from severe specificity collapse, with PL and PI yielding FPRs of 0.572 and 0.623, respectively. This demonstrates that SECT can operate safely at clinical thresholds without inundating radiologists with false positives.

To confirm that SECT resolves the primary failure mode of standard CNNs, we isolated the false positive evaluation strictly to bifurcation mimics (Frangi negatives). Under the optimal classification threshold, SECT successfully suppressed the bifurcation-specific FPR to 0.108 (Figure \ref{fig:roc2}, Panel 4). PI and PL, which lack the directional encoding necessary to capture asymmetric dome geometry, exhibited error rates nearly five times higher (0.516 and 0.517, respectively).

These results provide conclusive evidence that the performance gains of SECT are not merely aggregate improvements, but stem directly from its ability to resolve the specific geometric ambiguity between saccular aneurysms and branching vessels.

\end{document}